\documentclass{article}

\PassOptionsToPackage{numbers, compress, sort}{natbib}

\usepackage[preprint]{nips_2018}

\usepackage[utf8]{inputenc} 
\usepackage[T1]{fontenc}    
\usepackage{hyperref}       
\usepackage{url}            
\usepackage{booktabs}       
\usepackage{amsfonts}       
\usepackage{nicefrac}       
\usepackage{microtype}      
\usepackage{xcolor}

\usepackage{microtype}
\usepackage{graphicx}
\usepackage{subfigure}
\usepackage{booktabs} 
\usepackage{tikz}
\usetikzlibrary{matrix}
\usepackage{amsmath}
\usepackage{float}
\usepackage{comment}

\usepackage[boxed]{algorithm2e}
\usepackage{wrapfig}
\RequirePackage{authblk}

\usepackage[symbol]{footmisc}

\makeatletter

\g@addto@macro\normalsize{%
  \setlength\abovedisplayskip{1pt}
  \setlength\belowdisplayskip{1pt}
  \setlength\abovedisplayshortskip{1pt}
  \setlength\belowdisplayshortskip{1pt}
}

\title{Learning to Play With Intrinsically-Motivated, Self-Aware Agents}

\author[ ]{Nick Haber$^{1, 2, 3,}\footnote[1]{}$}
\author[ ]{Damian Mrowca$^{4,}\footnote[1]{}$}
\author[ ]{Stephanie Wang$^4$}
\author[ ]{Li Fei-Fei$^4$}
\author[ ]{\\Daniel L. K. Yamins$^{1, 4, 5}$}

\affil[ ]{Departments of Psychology$^{1}$, Pediatrics$^2$,  Biomedical Data Science$^3$, Computer Science$^4$, and Wu Tsai Neurosciences Institute$^5$, Stanford, CA 94305\vspace{0.15cm}}

\affil[ ]{\texttt{\{\href{mailto:nhaber@stanford.edu}{nhaber}, \href{mailto:mrowca@stanford.edu}{mrowca}\}@stanford.edu}\vspace{-.6cm}}

\begin{document}

\maketitle

\footnotetext[1]{Equal contribution}

\begin{abstract}
Infants are experts at playing, with an amazing ability to generate novel structured behaviors in unstructured environments that lack clear extrinsic reward signals. We seek to mathematically formalize these abilities using a neural network that implements curiosity-driven intrinsic motivation.  Using a simple but ecologically naturalistic simulated environment in which an agent can move and interact with objects it sees, we propose a ``world-model'' network that learns to predict the dynamic consequences of the agent's actions.  Simultaneously, we train a separate explicit ``self-model'' that allows the agent to track the error map of its world-model. It then uses the self-model to adversarially challenge the developing world-model. We demonstrate that this policy causes the agent to explore novel and informative interactions with its environment, leading to the generation of a spectrum of complex behaviors, including ego-motion prediction, object attention, and object gathering.  Moreover, the world-model that the agent learns supports improved performance on object dynamics prediction, detection, localization and recognition tasks.  Taken together, our results are initial steps toward creating flexible autonomous agents that self-supervise in realistic physical environments.
\end{abstract}

\vspace{-.3cm}
\section{Introduction}
\label{introduction}
\vspace{-.2cm}

Truly autonomous artificial agents must be able to discover useful behaviors in complex environments without having humans present to constantly pre-specify tasks and rewards. 
This ability is beyond that of today's most advanced autonomous robots. 
In contrast, human infants exhibit a wide range of interesting, apparently spontaneous, visuo-motor behaviors --- including navigating their environment, seeking out and attending to novel objects, and engaging physically with these objects in novel and surprising ways \citep{fantz_visualexperienceininfants, twomey_curiositybasedlearning, hurley2010_influenceofpets, hurley2015_petexposure, goupil_infantsask, begus_infantslearnwhattheywant, gopnik_scientistincrib}.
In short, young children are excellent at playing --- ``scientists in the crib'' \citep{gopnik_scientistincrib} who create, intentionally, events that are new, informative, and exciting to them \citep{sokolov1963perception, fantz_visualexperienceininfants}. 
Aside from being fun, play behaviors are an active learning process \citep{settles2011_active}, driving self-supervised learning of representations underlying sensory judgments and motor planning \citep{kidd2012_goldilocks, goupil_infantsask, begus_infantslearnwhattheywant}.

\begin{wrapfigure}{R}{0.5\textwidth}
\begin{center}
\includegraphics[width=0.48\textwidth]{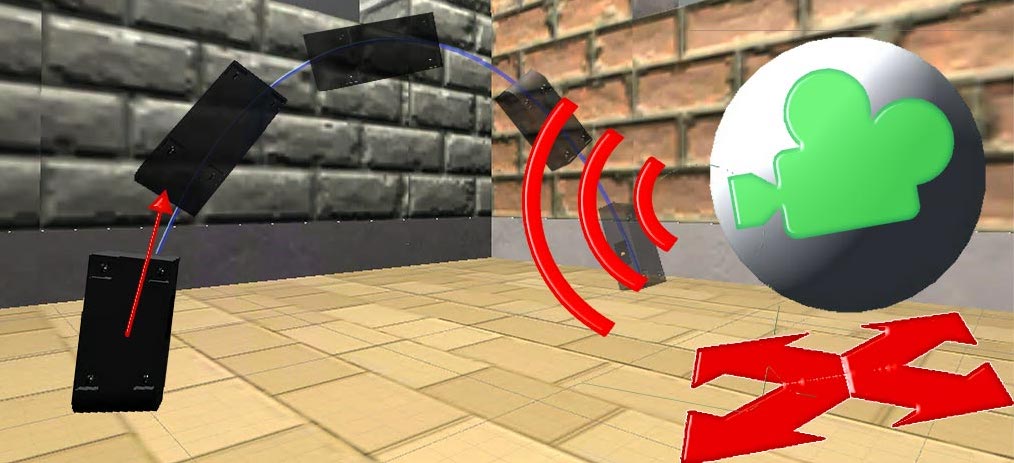}
\end{center}
\vspace{-0.3cm}
\caption{\emph{3D Physical Environment.} The agent can move around, apply forces to visible objects in close proximity, and receive visual input.}
\label{fig:setup}
\vskip -0.1in
\end{wrapfigure}


But how can we use these observations on infant play to improve artificial intelligence?
AI theorists have long realized that playful behavior in the absence of rewards can be mathematically formalized via loss functions encoding intrinsic reward signals, in which an agent chooses actions that result in novel but predictable states that maximize its learning \citep{schmidhuber_formaltheoryoffun}.
These ideas rely on a virtuous cycle in which the agent actively self-curricularizes as it pushes the boundaries of what its world-model-prediction systems can achieve.
As world-modeling capacity improves, what used to be novel becomes old hat, and the cycle starts again.

Here, we build on these ideas using the tools of deep reinforcement learning to create an artificial agent that learns to play.
We construct a simulated physical environment inspired by infant play rooms, in which an agent can swivel its head, move around, and physically act on nearby visible objects (Fig. 1).
Akin to challenging video game tasks \citep{kulkarni_hierarchical}, informative interactions in this environment are possible, but sparse unless actively sought by the agent.
However, unlike most video game or constrained robotics environments, there is no extrinsic goal to constrain the agent's action policy.  The agent has to learn about its world, and what is interesting in it, for itself.

In this environment, we describe a neural network architecture with two interacting components, a \emph{world-model} and a \emph{self-model}, which are learned simultaneously.  The world-model seeks to predict the consequences of agent’s actions, either through forward or inverse dynamics estimation.
The self-model learns explicitly to predict the errors of the world-model. 
The agent then uses the self-model to choose actions that it believes will adversarially challenge the current state of its world-model.  

Our core result is the demonstration that this intrinsically-motived self-aware architecture stably engages in a virtuous reinforcement learning cycle, spontaneously discovering highly nontrivial cognitive behaviors --- first understanding and controlling self-generated motion of the agent (``ego-motion''), and then selectively paying attention to, and eventually organizing, objects.
This learning occurs through an emergent active self-supervised process in which new capacities arise at distinct ``developmental milestones'' like those in human infants. 
Crucially, it also learns visual encodings with substantially improved transfer to key visual scene understanding tasks such as object detection, localization, and recognition and learns to predict physical dynamics better than a number of strong baselines.  
This is to our knowledge the first demonstration of the efficacy of active learning of a deep visual encoding for a complex three-dimensional environment in a purely self-supervised setting.
Our results are steps toward mathematically well-motivated, flexible autonomous agents that use intrinsic motivation to learn about and spontaneously generate useful behaviors for real-world physical environments.

\textbf{Related Work}
Our work connects to a variety of existing ideas in self-supervision, active learning, and deep reinforcement learning. Visual learning can be achieved through self-supervised auxiliary tasks including semantic segmentation \citep{selfsupervision_semanticsegmentation}, pose estimation \citep{selfsupervision_pose}, solving jigsaw puzzles \citep{jigsaw_puzzles}, colorization \citep{colorization}, and rotation \citep{rotnet}. Self-supervision on videos frame prediction \citep{kalchbrenner2016video} is also promising, but faces the challenge that most sequences in recorded videos are ``boring'', with little interesting dynamics occurring from one frame to the next. 

In order to encourage interesting events to happen, it is useful for an agent to have the capacity to select the data that it sees in training. In active learning, an agent seeks to learn a supervised task using minimal labeled data \citep{settles2011_active, giladbachrach_qbcreal_active}. Recent methods obtain diversified sets of hard examples \citep{elhamifar2013convex, sener2017active}, or use confidence-based heuristics to determine when to query for more labels \citep{wang2016cost}. Beyond selection of examples from a pre-determined set, recent work in robotics \citep{agrawal2016learning,popov2017data,finn2017deep, ebert2017self} study learning tasks with interactive visuo-motor setups such as robotic arms. The results are promising, but largely use random policies to generate training data without biasing the robot to explore in a structured way.

Intrinsic and extrinsic reward structures have been used to learn generic ``skills'' for a variety of tasks \citep{chentanez2005intrinsically, singh2010intrinsically, machado2017laplacian}. 
\citet{houthooft_vime} demonstrated that reasonable exploration-exploitation trade-offs can be achieved by intrinsic reward terms formulated as information gain. 
\citet{frank2014curiosity} use information gain maximization to implement artificial curiosity on a humanoid robot. 
\citet{kulkarni_hierarchical} combine intrinsic motivation with hierarchical action-value functions operating at different temporal scales, for goal-driven deep reinforcement learning.   
\citet{achiam2017surprise} formulate surprise for intrinsic motivation as the KL-divergence of the true transition probabilities from learned model probabilities. 
\citet{held2017automatic} use a generator network, which is optimized using adversarial training to produce tasks that are always at the appropriate level of difficulty for an agent, to automatically produce a curriculum of navigation tasks to learn. 
\citet{jaderberg2016reinforcement} show that target tasks can be improved by using auxiliary intrinsic rewards.

Oudeyer and colleagues \citep{oudeyer2007intrinsic, oudeyer2016evolution, gottlieb2013information} have explored formalizations of curiosity as maximizing prediction-ability change, showing the emergence of interesting realistic cognitive behaviors from simple intrinsic motivations.  Unlike this work, we use deep neural networks to learn the world-model and generate action choices, and co-train the world-model and self-model, rather than pre-training the world-model on a separate prediction task and then freezing it before instituting the curious exploration policy.  \citet{berkeley_mario} uses curiosity to antagonize a future prediction signal in the latent space of a inverse dynamics prediction task to improve learning in video games, showing that intrinsic motivation leads to faster floor-plan exploration in a 2D game environment.  
Our work differs in using a physically realistic three-dimensional environment and shows how intrinsic motivation can lead to substantially more sophisticated agent-object behavior generation (the ``playing'').  Underlying the difference between our technical approach is our introduction of a self-model network, representing the agent's awareness of its own internal state. 
This difference can be viewed in RL terms as the use of a more explicit model-based architecture in place of a model-free setup. 

Unlike previous work, we show the learned representation transfers to improved performance on analogs of real-world visual tasks, such as object localization and recognition.
To our knowledge, a self-supervised setup in which an explicitly self-modeling agent uses intrinsic motivation to learn about and restructure its environment has not been explored prior to this work.

\vspace{-.3cm}
\section{Environment and Architecture} \label{sec:env_and_architecture}
\vspace{-.2cm}

\begin{figure*}
\begin{center}
\includegraphics[width=\textwidth]{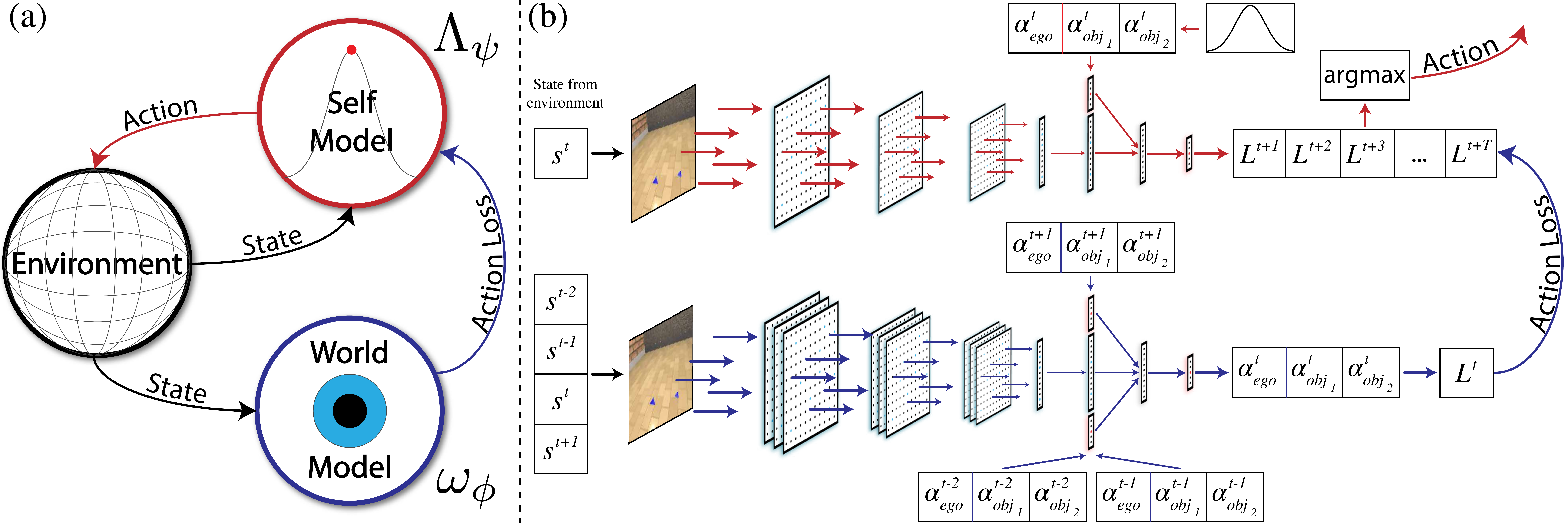}
\end{center}
\vspace{-0.3cm}
\caption{\textbf{Intrinsically-motivated self-aware agent architecture.} The world-model (blue) solves a dynamics prediction problem. Simultaneously a self-model (red) seeks to predict the world-model's loss. Actions are chosen to antagonize the world-model, leading to novel and surprising events in the environment (black). (a) Environment-agent loop. (b) Agent information flow.}
\label{fig:model}
\vskip -0.35in
\end{figure*}

\textbf{Interactive Physical Environment.}
Our agent is situated in a physically realistic simulated {\em environment} (black in Fig. \ref{fig:model}) built in Unity 3D (Fig. \ref{fig:setup}). 
Objects in the environment interact according to Newtonian physics as simulated by the PhysX engine \citep{boeing2007evaluation}. 
The agent's avatar is a sphere that swivels in place, moves around, and receives RGB images from a forward-facing camera (as in Fig. \ref{fig:setup}). 
The agent can apply forces and torques in all three dimensions to any objects that are both in view and within a fixed {\em maximum interaction distance} $\delta$ of the agent's position. We say that such an object is in a {\em play state}, and that a state with such an object is a {\em play state}.
Although the floor and walls of the environment are static, the agent and objects can collide with them.
The agent's action space is a subset of $\mathbb{R}^{2+6N}$. The first $2$ dimensions specify ego-motion, restricting agent movement to forward/backward motion $v_{fwd}$ and horizontal planar rotation $v_{\theta}$, while the remaining $6N$ dimensions specify the forces $f_x, f_y, f_z$ and torques $\tau_x, \tau_y, \tau_z$ applied to $N$ objects sorted from the lower-leftmost to the upper-rightmost object relative to the agent's field of view. 
All coordinates are bounded by constants and normalized to be within $[-1, 1]$ for input into models and losses.
In this setup, both the observation space (images from the 3d rendering) and action space (ego-motion and object force application) are continuous and high-dimensional, complicating the challenges of learning the visual encoding and action policy.

\textbf{Agent Architecture.}
Our agent consists of two simultaneously-learned components: a \emph{world-model} and a \emph{self-model} (Fig. \ref{fig:model}). 
The world-model seeks to solve one or more \emph{dynamics prediction problems} based on inputs from the environment. The self-model seeks to estimate the world-model's losses for several timesteps into the future, as a function both of visual input and of potential agent actions.  
An action choice policy based on the self-model's output chooses actions that are ``interesting'' to the world-model. In this work, we choose perhaps the simplest such motivational mechanism, using policies that try to maximize the world-model's loss.
In part as a review of the key issues of prediction error-based curiosity \cite{berkeley_mario,schmidhuber_formaltheoryoffun,oudeyer2007intrinsic, oudeyer2016evolution}, we now formalize these ideas mathematically.

\textbf{\textit{World-Model:}}
At the core of our architecture is the world-model --- e.g. the neural network that attempts to learn how dynamics in the agent's environment work, especially in reaction to the agent's own actions.  
Finding the right dynamics prediction problem(s) to set as the agent's world-modeling goal is a nontrivial challenge.  

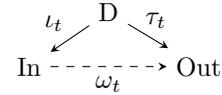
\begin{wrapfigure}{R}{0.4\textwidth} 
\vspace{-3ex}
\begin{center}
\begin{tikzpicture}
  \matrix (m) [matrix of math nodes,row sep=0.7em,column sep=1.5em,minimum width=1em]
  {
    & \operatorname{D} & \\
        \operatorname{In} & & \operatorname{Out}\\
  };
    \path[-stealth]
    (m-1-2) edge node [xshift=-0.6em, yshift=0.6em] {$\iota_t$} (m-2-1)
        edge node [xshift=0.3em, yshift=0.6em] {$\tau_t$} (m-2-3)
  (m-2-1) edge [dashed] node [below] {$\omega_t$} (m-2-3)
  ;
\end{tikzpicture}
\end{center}
\vspace{-3ex}
\caption{Diagramming dynamics prediction problems.}
\vspace{-3ex}
\label{fig:triangle}
\end{wrapfigure}

Consider a partially observable Markov Decision Process (POMDP) with state $s_t$, observation $o_t$, and action $a_t$.  In our agent's situation, $s_t$ is the complete information of object positions, extents, and velocities at time $t$; $o_t$ is the images rendered by the agent-mounted camera; and $a_t$ is the agent's applied ego-motion, forces and torques vector.  
The rules of physics are the dynamics which generate $s_{t+1}$ from $s_t$ and $a_t$. Agents make decisions about what action to take at each time, accumulating histories of observations and actions.
Informally, a {\em dynamics prediction problem} is a pairing of complementary subsets of data --- ``inputs'' and ``outputs'' --- generated from the history.
The goal of the agent is to learn a map from inputs to outputs. 
More precisely, adopting the notation $o_{t_1 : t_2} = (o_{t_1}, o_{t_1 + 1}, \ldots o_{t_2})$ and similarly for actions and states, we let D (the ``data'') be fixed-time-length segments of history $\{d_t = (o_{t - b : t + f}, a_{t - b : t + f}) \ | \ t = 1, 2, 3\ldots\}$. A dynamics prediction problem (Figure~\ref{fig:triangle}) is then defined by specifying (possibly time-varying) maps $\iota_t: \operatorname{D} \rightarrow \operatorname{In}$ and $\tau_t: \operatorname{D} \rightarrow \operatorname{Out}$ for some specified input and output spaces In and Out, forming a triangular diagram.

Also given as part of the dynamics prediction problem is a loss $L$ for comparing ground-truth versus estimated outputs.
The agent's \emph{world-model} at time $t$ is a map $\omega_{\theta_t}:~\operatorname{In}~\rightarrow~\operatorname{Out}$
whose parameters are updated by stochastic gradient descent in order to lower $L$.
In words, the agent's world-model (blue in Fig.~\ref{fig:model}) tries to learn to reconstruct the true-value from the input datum.
Note that batches of data on which this update occurs are not drawn from any fixed distribution since they come from the history of an agent as it executes its policy, and hence this learning process does not correspond to a traditional statistical learning optimization.

Since we are focused on agents learning from an environment without external input, the maps $\iota$ and $\tau$ should in general be easy for the agent to estimate at low cost from its ``sense data'' --- what is sometimes called self-supervision \citep{selfsupervision_semanticsegmentation, selfsupervision_pose, jigsaw_puzzles, colorization, rotnet, kalchbrenner2016video}.
For example, perhaps the most natural dynamics problem to assign to the agent as the goal of its world-model is \emph{forward dynamics prediction}, with input $(o_{t-b:t}, a_{t-b:t+f-1})$ and true-value $o_{t+1:t+f}$.
In words, the agent is trying to predict the next (several) observation(s) given past observations and a sequence (past and present) of actions.
In 3-D physical domains such as ours, the outputs correspond to $f$ bitmap image arrays of future frames, and the loss function $L_F$ may be $\ell_2$ loss on pixels or some discretization thereof.
Despite recent progress on the frame prediction problem \citep{kalchbrenner2016video, finn2017deep}, it remains quite challenging, in part because the dimensionality of the true-value space is so large.

In practice, it can be substantially easier to solve \emph{inverse dynamics prediction}, with input $(o_{t - b:t + f}, a_{t - b: t  - 1}, a_{t + 1 : t + f - 1})$ and true-value $a_t$.
In other words, the agent is trying to ``post-dict'' the action needed to have generated the observed sequence of observations, given knowledge of its past and future actions.
Here, the loss function $L_{ID}$ is computed on (what is generally the comparitively low-dimensional) action space, a problem that has proven tractable \citep{baranes2013active, agrawal2016learning}.

One major concern in intrinsic motivation, in particular when the agent's policy attempts to maximize the world-model's loss, is when the dynamics prediction problem is inherently unpredictable. This is sometimes referred to (perhaps in less generality than in what we proceed to define) as the {\em white-noise problem} \citet{berkeley_mario,schmidhuber_formaltheoryoffun}. In cases where the agent's policy attempts to maximize the world-model's loss, the agent is motivated to fixate on the unlearnable. Within the above framework, this problem manifests in that there is no requirement that $\iota_t$ and $\tau_t$ actually induce a well-defined mapping $\operatorname{In} \rightarrow \operatorname{Out}$ that makes the diagram above commute. We refer to the existence of policies for which there are obstructions to such a commuting diagram, with nonzero probability, as \emph{degeneracy} in the dynamics prediction problem. In fact, the inverse dynamics problem can suffer from substantial degeneracy. Consider the case of an agent pressing an object straight into the ground: no matter what the downward force is, the object does not move, so the vision and action input information is insufficient to determine the true-value.

To avoid both of these pixel space and degeneracy difficulties, one can instead try forward dynamics prediction, but in a latent space --- for example, the latent space determined by an encoder for inverse dynamics problem \citep{berkeley_mario}.
In this case, we begin with a system solving the inverse dynamics prediction problem and assume that its parametrization of world-model factors into a composition $\omega^{ID}_{\theta_t} = d^{ID}_{\beta_t} \circ e^{ID}_{\alpha_t}$ where $\alpha_t$ and $\beta_t$ are non-overlapping sets of parameters.  
We call $e^{ID}_t = e^{ID}_{\alpha_t}$ the \emph{encoding} and the range of $e^{ID}_t$ the \emph{latent space} $\mathcal{L}$ of the ID problem. On top, we define (time-varying) $\iota^{LF}_{t}, \tau^{LF}_{t}$ as the 1-time-step future prediction problem on trajectories in $\mathcal{L}$ given by the time-varying encoding, i.e. by $\iota^{LF}_{t}(d_t) = (e^{ID}_t(o_{t - b : t}), a_{t-b:0})$ and 
$\tau^{LF}_t(d_t) = e^{ID}_t(o_{t+1})$. The problem is then supervised by $\ell_2$ loss.
The inverse-prediction world-model $\omega^{ID}$ and latent-space world-model $\omega^{LF}$ evolve simultaneously.
If $\mathcal{L}$ is sufficiently low dimensional, this may be a good compromise task that represents only ``essential'' features for prediction.

In this work, we explore both inverse dynamics and latent space future prediction tasks.

\textbf{\textit{Explicit Self-Model:}}
In the strategy outlined above, the agent's action policy goal is to antagonize its world-model.  
If the agent explicitly predicts its own world-model loss $L_{\omega_t}$ incurred at future timesteps as a function of visual input and current action, a simple antagonistic policy could simply seek to maximize $L_{\omega_t}$ over some number of future timesteps.
Embodying this idea, the self-model $\Lambda$ (red in Fig.~\ref{fig:model}) is given $o_{t-1 : t}$ and a proposed next action $a$ and predicts
$\Lambda_{o_{t-1:t}}(a) = (p_1(c \ | \ o_{t-1:t}, a)\ldots p_T(c \ | \ o_{t-1:t}, a)),$
where $p_i(c  \ | \ o_{t-1:t}, a)$ is the probability that the loss incurred by the world-model at time $t+i$ will equal $c$.
For convenience of optimization, we discretize the losses into loss bins $C$, so that each $p_i \in \mathcal P(C)$ is a probability distribution over discrete classes $c \in C$. 
$\Lambda_{o_{t-1:t}}(a)$ is penalized with a softmax cross-entropy loss for each $i$ and averaged over $i \in {1, \ldots, T}$. 
All future losses aside from the first one depend on future actions taken, and the self-model hence needs to predict in expectation over policy. 
Each $p_i(c  \ | \ o_{t-1:t}, a)$ can be interpreted as a \emph{map} over action space which turns out to be useful for intuitively visualizing what strategy the agent is taking in any given situation (see Fig \ref{fig:lossmap_examples}).

\textbf{\textit{Adversarial Action Choice Policy:}}
The self-model provides, given $o_{t-1:t}$ and a proposed next action $a$, $T$ probability distributions $p_i$. 
The agent uses a simple mechanism to convert this data to an action choice.
To summarize loss map predictions over times $t \in \{1, \ldots T\}$, we add expectation values:
$$\sigma(a)[o_{t-1:t}] = \sum_i \sum_{c \in C} c \cdot p_i(c).$$
The agent's action policy is then given by sampling with respect to a Boltzmann distribution $\pi(a \ | \ o_{t-1:t}) \sim \exp(\beta \sigma(\Lambda_{o_{t-1:t}}(a)))$ with fixed hyperparameter $\beta$.

\begin{figure*}
\begin{center}
  \includegraphics[width=.9\linewidth]{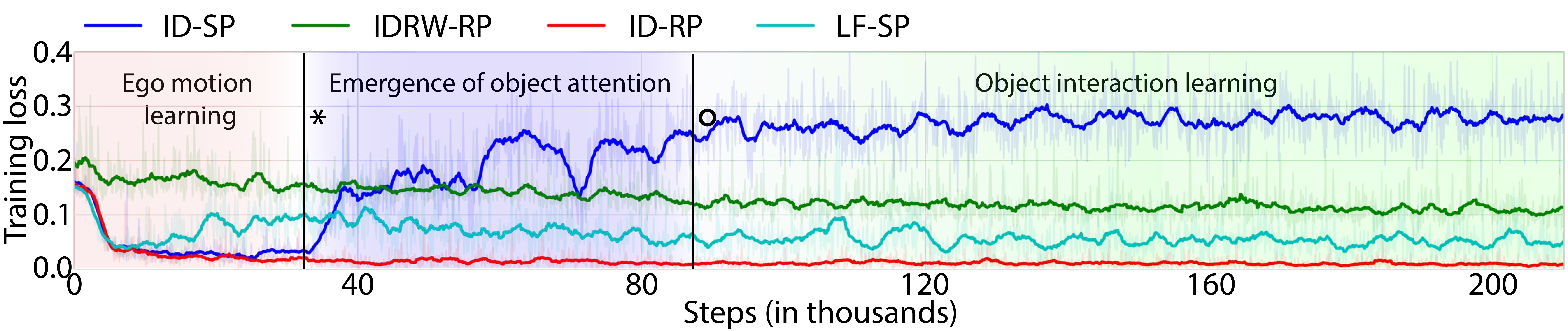}
    
    (a) World-model training loss
    
  \begin{tabular}[c]{@{\hspace{-0.1pt}}c@{\hspace{10.0pt}}c@{\hspace{10.0pt}}c}
    \includegraphics[width=0.32\linewidth]{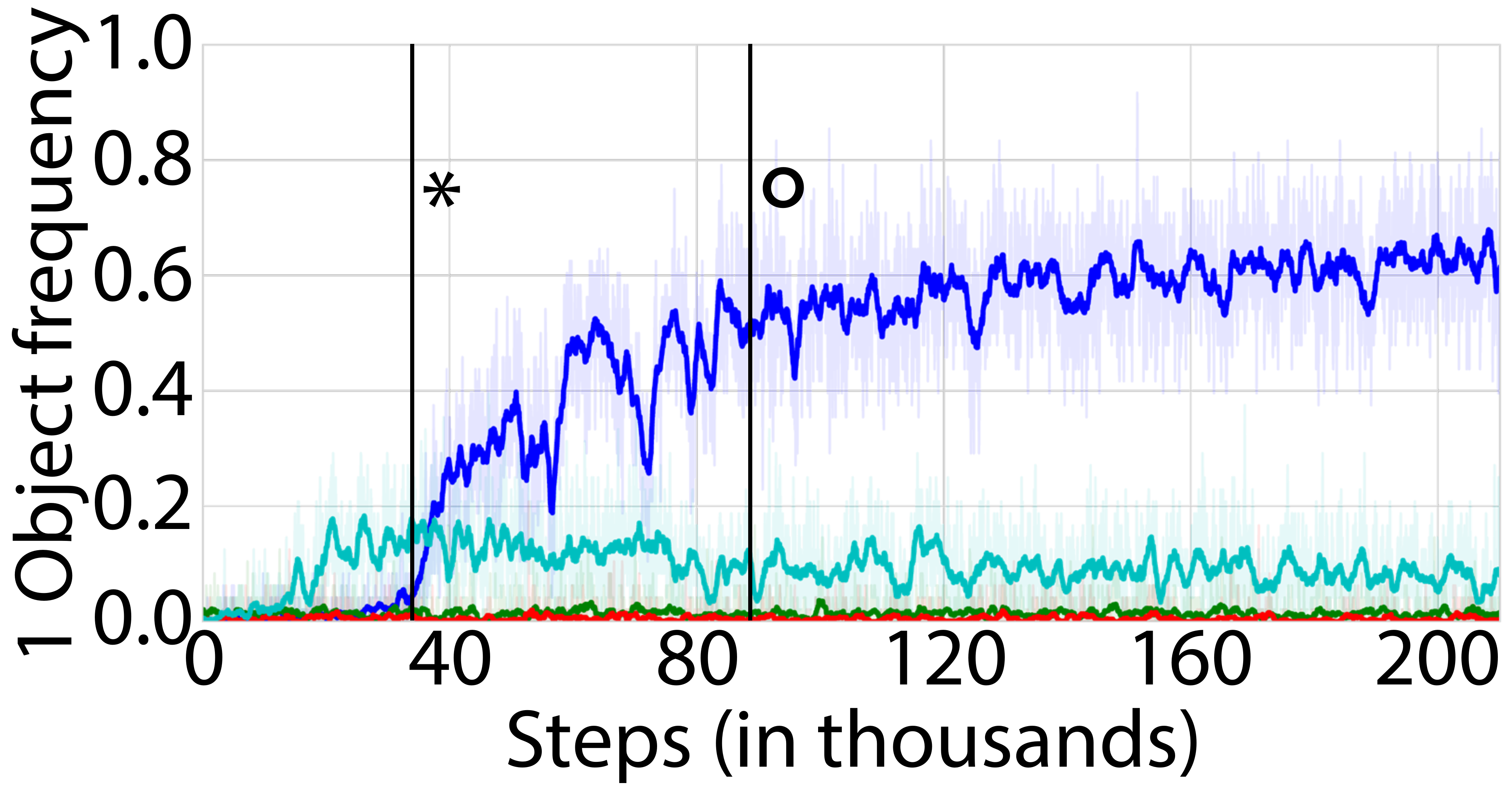}
        &
        \includegraphics[width=0.3\linewidth]{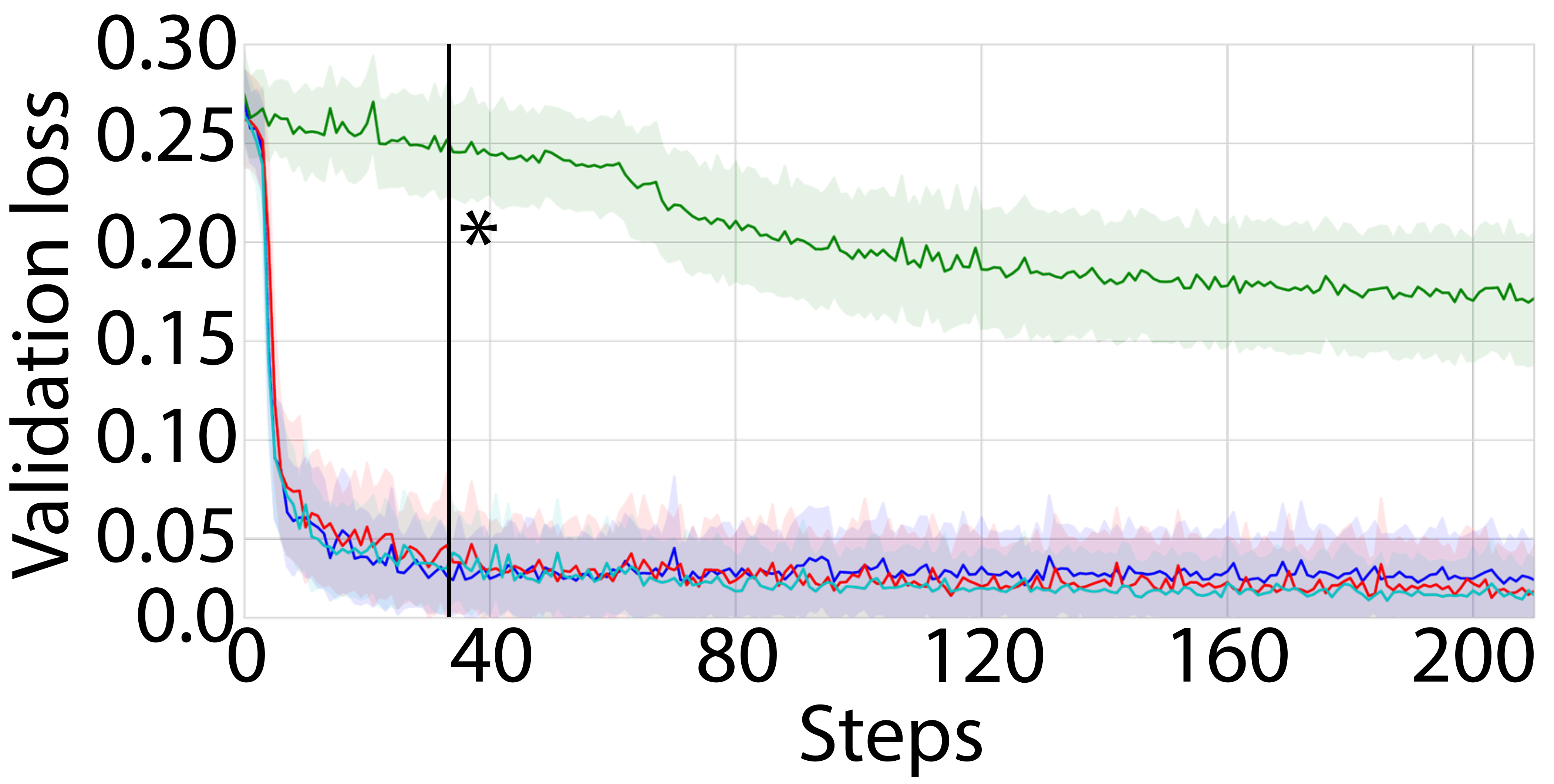} 
        &
        \includegraphics[width=0.3\linewidth]{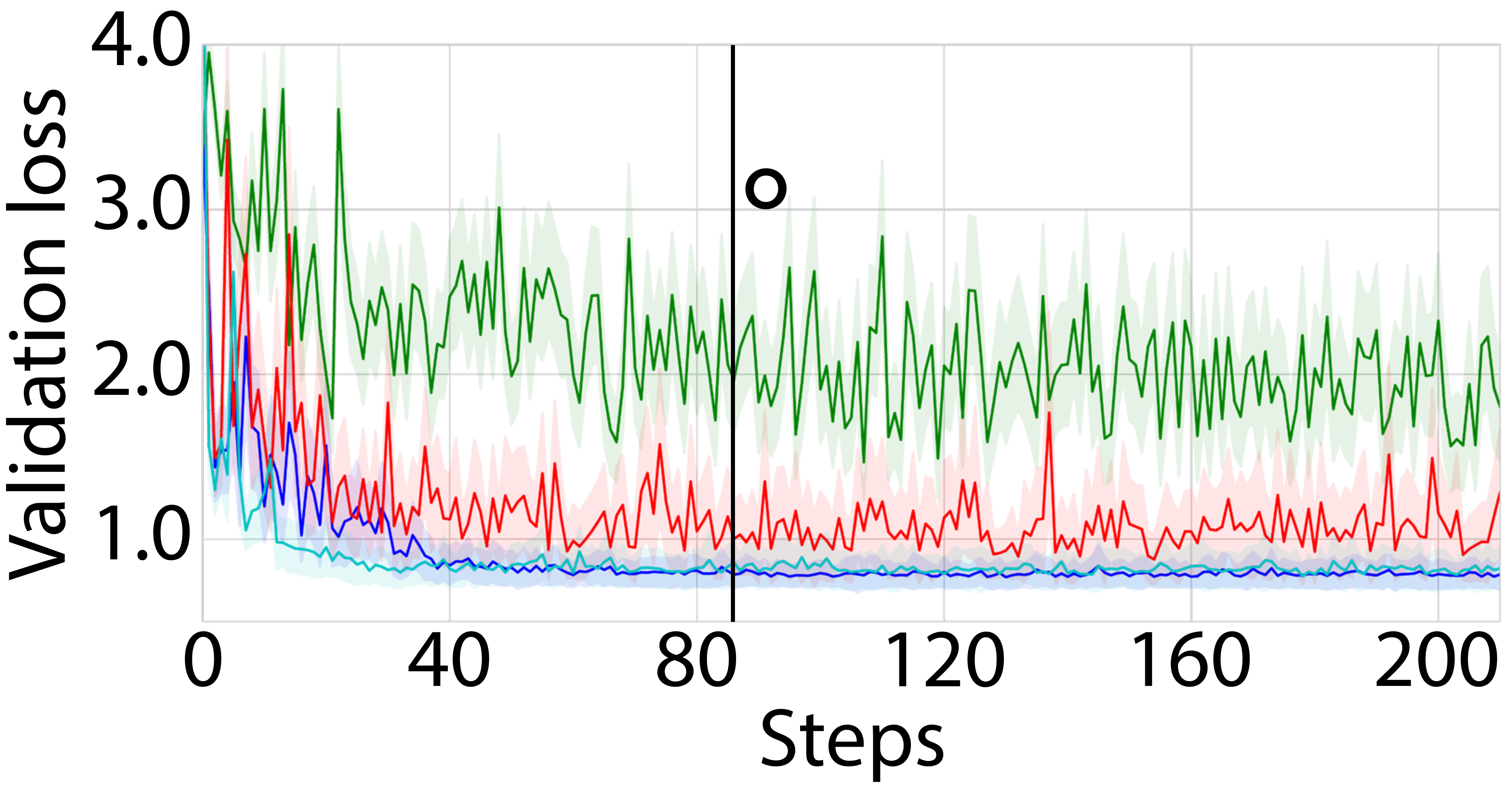}
        \\
        (b) Object frequency
        &
        (c) Easy data (ego-motion)
        &
        (d) Hard data (object play)
        \\
  \end{tabular}
\end{center}
\vspace{-0.3cm}
\caption{\textbf{Single-object experiments.} (a) World-model training loss. (b) Percentage of frames in which an object is present. (c) World-model test-set loss on ``easy'' ego-motion-only data, with no objects present.   (d) World-model test-set loss on ``hard'' validation data, with object present, where agent must solve object physics prediction. Validation datasets are evaluated every 1000 batch update steps. Lighter curves represent unsmoothed batch-mean values --- min and max for validation.}
\label{fig:1obj_experiments}
\vskip -0.2in
\end{figure*}

\textbf{\textit{Architectures and Losses:}}
We use convolutional neural networks to learn both world-models $\omega_\theta$ and self-models $\Lambda_\psi$.
In the experiments described below, these have an encoding structure with a common architecture involving twelve convolutional layers, two-stride max pools every other layer, and one fully-connected layer, to encode observations into a lower-dimensional latent space, with shared weights across time.
For the inverse dynamics task, the top encoding layer of the network is combined with actions $\{a_{t'}\ |\  t' \neq t\}$, fed into a two-layer fully-connected network, on top of which a softmax classifier is used to predict action $a_t$. 
For the latent space future prediction task, the top convolutional layer of $\omega^{ID}_{\theta_{ID}}$ is used as the latent space $\mathcal{L}$, and the latent model $\omega^{LF}_{\theta_{LF}}$ is parametrized by a fully-connected network that receives, in addition to past encoded images, past actions. In the ID-only case (ID-SP), we optimize $\min_{\theta_{ID}} L_{ID} + \min_{\psi} L_{\Lambda, ID}$. In the LF case (LF-SP), we optimize $\min_{\theta_{ID}} L_{ID} + \min_{\theta_{LF}} L_{LF} + \min_{\psi} L_{\Lambda, LF}$. See supplementary for details.

\vspace{-.3cm}
\section{Experiments}
\vspace{-.2cm}

\label{sec:experiments}
We randomly place the agent in a square 10 by 10 meter room, together with up to two other objects with which the agent can interact, setting the maximum interaction distance $\delta$ to 2 meters. The objects are drawn from a set of 16 distinct geometric shapes, e.g. cones, cylinders, cuboids, pyramids, and spheroids of various aspect ratios. Training is performed using 16 asynchronous simulation instances~\citep{mnih_asynchronousrl}, with different seeds and objects. The scene is reinitialized periodically, with time of reset randomly chosen between $2^{13}$ to $2^{15}$ steps. Each simulation maintains a data buffer of 250 timesteps to ensure stable training~\citep{Lin_rlmethods_experiencereplay}. For model updates two examples are randomly sampled from each of the 16 simulation buffers to form a batch of size 32.  Gradient updates are performed using the Adam algorithm \citep{kingma2014adam}, with an initial learning rate of $0.0001$. See the supplement for tests of the stability of all results to variations in interaction radius $\delta$, room size, and agent speed, as well as per-object-type behavioral breakdowns.

For each experiment, we evaluate the agents' abilities with three types of metrics.  We first measure the \emph{(i) spontaneous emergence of novel behaviors}, involving the appearance of highly structured but non-preprogrammed events such as the agent attending to and acting upon objects (rather than just performing mere self-motion), engaging in directed navigation trajectories, or causing interactions between multiple objects. Finding such emergent behaviors indicates that the curiosity-driven policies generate qualitatively novel scenarios in which the agent can push the boundaries of its world-model. For each agent type, we also evaluate \emph{(ii) improvements in dynamic task prediction} in the agents' world-models, on challenging held-out validation data constructed to test learning about both ego-motion dynamics and object physical interactions. Finding such improvements indicates that the data gleaned from the novel scenarios uncovered by intrinsic motivation actually does improve the agents' world-modeling capacities.  Finally, we also evaluate \emph{(iii) task transfer}, the ability of the visual encoding features learned by the curious agents to serve as a general basis for other useful visual tasks, such as object recognition and detection.

\textit{Control models:} In addition to the two curious agents, we study several ablated models as controls. \emph{ID-RP} is an ablation of ID-SP in which the world-model trains but the agent executes a random policy, used to demonstrate the difference an active policy makes in world-model performance and encoding. \emph{IDRW-SP} is an ablation of ID-SP in which the policy is executed as above but with the encoding portion of the world-model frozen with random weights.  This control measures the importance of having the action policy inform the deep internal layers of the world model network. \emph{IDRW-RP} combines both ablations.


\vspace{-.2cm}
\subsection{Emergent behaviors}
\vspace{-.2cm}

Using metrics inspired by the developmental psychology literature, we quantify the appearance of novel structured behaviors, including attention to and acting on objects, navigation and planning, and ability to interact with multiple objects. In addition to sharp stage-like transitions in world-model loss and self-model evaluations, to quantify these behaviors we measure play state frequency and (in the case of multiple objects) the average distance between the agent and objects. We compute these quantities by averaging play state count and distance between objects, respectively, over the three simulation steps per batch update. Quantities presented below are aggregates over all 16 simulation instances unless otherwise specified.

\textbf{Object attention}.
Fig. \ref{fig:1obj_experiments}a shows the total training loss curves of the ID-SP, LF-SP models and baselines. Upon initialization, all agents start with behaviors indistinguishable from the random policy, choosing largely self-motion actions and rarely interacting with objects. For learned-weight agents, an initial loss decrease occurs due to learning of ego-motion, as seen in Fig.~\ref{fig:1obj_experiments}a.  For the curious agents, this initial phase is robustly succeeded by a second phase in which loss increases.  As shown in Fig.~\ref{fig:1obj_experiments}b, this loss increase corresponds to the emergence of object attention, in which the agent dramatically increases the play state frequency. As seen by comparing Fig.~\ref{fig:1obj_experiments}c-d, object interactions are much harder to predict than simple ego-motions, and thus are enriched by the curious policy: for the ID-SP agent, object interactions increase to about $60\ \%$ of all frames.  In comparison, frequency of object interaction increases much less or not at all for control policy agents.

\begin{figure*}
\begin{center}  
  \begin{tabular}[c]{c@{\hspace{3.0pt}}c@{\hspace{3.0pt}}c@{\hspace{3.0pt}}c@{\hspace{3.0pt}}c@{\hspace{3.0pt}}c@{\hspace{3.0pt}}c}
         \includegraphics[width=0.15\textwidth]{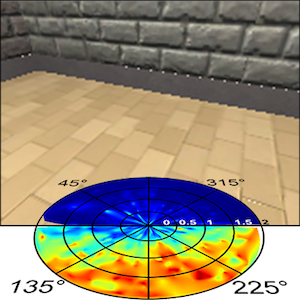}
         &
         \includegraphics[width=0.15\textwidth]{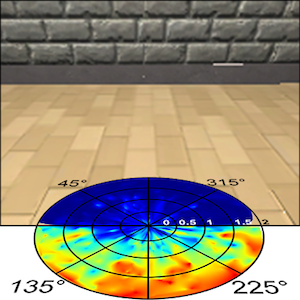}
         &
         \includegraphics[width=0.15\textwidth]{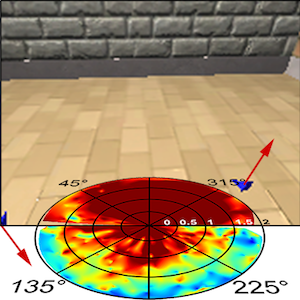}
         &
         \includegraphics[width=0.15\textwidth]{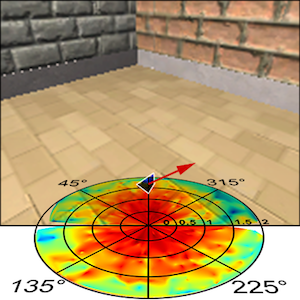}
         &
         \includegraphics[width=0.15\textwidth]{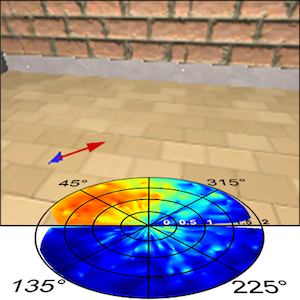}
         &
         \includegraphics[width=0.15\textwidth]{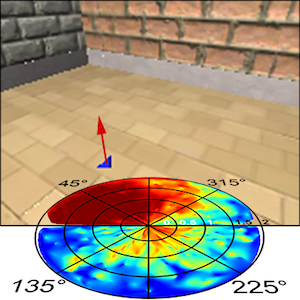}
         &
         \includegraphics[width=0.03\textwidth]{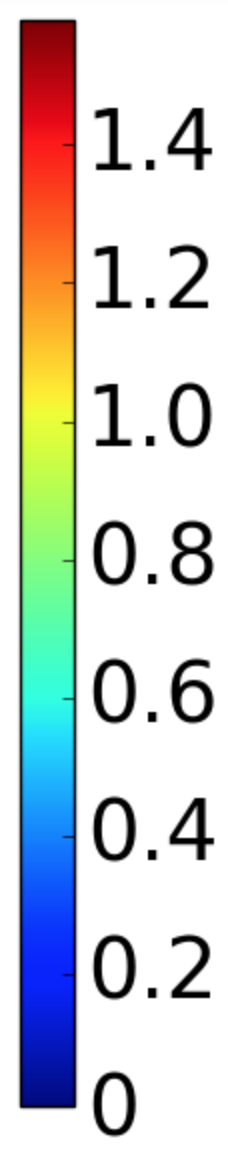}
         \\
         t & t+1 & t+2 & t+3 & t+4 & t+5 & \\
         \includegraphics[width=0.15\textwidth]{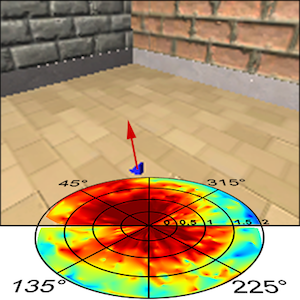}
         &
         \includegraphics[width=0.15\textwidth]{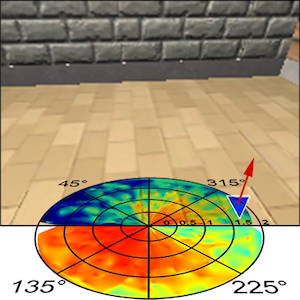}
         &
         \includegraphics[width=0.15\textwidth]{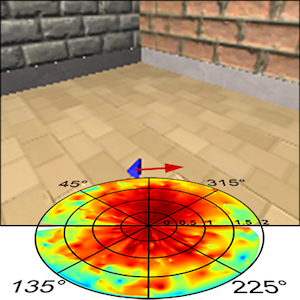}
         &
         \includegraphics[width=0.15\textwidth]{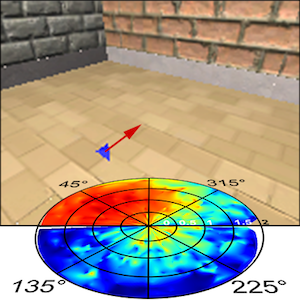}
         &
         \includegraphics[width=0.15\textwidth]{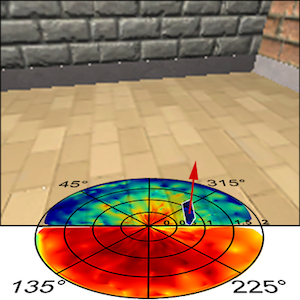}
         &
         \includegraphics[width=0.15\textwidth]{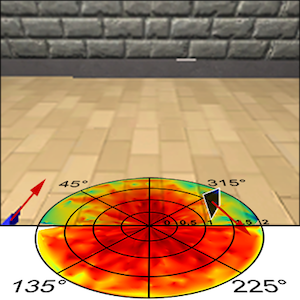}
         &
         \includegraphics[width=0.03\textwidth]{images2/ex6_cb.png}
         \\
         t+6 & t+7 & t+8 & t+9 & t+10 & t+11 & \\
  \end{tabular}
\end{center}
\vspace{-0.3cm}
\caption{\textbf{Navigation and planning behavior.} Example model roll-out for 12 consecutive timesteps. Red force vectors on the objects depict the actions predicted to maximize world-model loss. Ego-motion self-prediction maps are drawn at the center of the agents position. Red colors correspond to high and blue colors to low loss predictions. The agent starts without seeing an object and predicts higher loss if it turns around to explore for an object. The self-model predicts higher loss if the agent approaches a faraway object or turns towards a close object to keep it in view.}
\label{fig:lossmap_examples}
\vskip -0.2in
\end{figure*}

\textbf{Navigation and planning}.
The curiousity-driven agents also exhibit emergent navigation and planning abilities. In Fig. \ref{fig:lossmap_examples} we visualize ID-SP self-prediction maps projected onto the agent's position for the one-object setup. The maps are generated by uniformly sampling $1000$ actions $a$, evaluating $\Lambda_{o_{t-1:t}}(a)$ and applying a post-processing smoothing algorithm. We show an example sequence of 12 timesteps. The self-prediction maps show the agent predicting a higher loss (red) for actions moving it towards the object to reach a play state. As a result, the intrinsically-motivated agents learn to take actions to navigate closer to the object.  

\begin{figure*}
\begin{center}  
  \includegraphics[width=\linewidth]{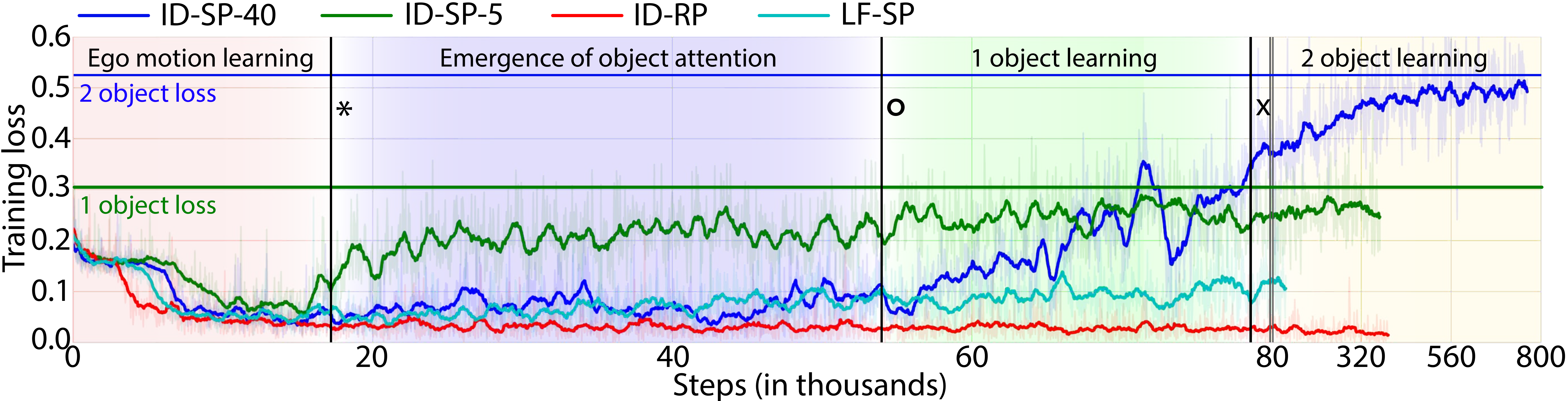}
    
    (a) World-model training loss
  \begin{tabular}[c]{@{\hspace{-0.1pt}}ccc}
    \includegraphics[width=0.3\linewidth]{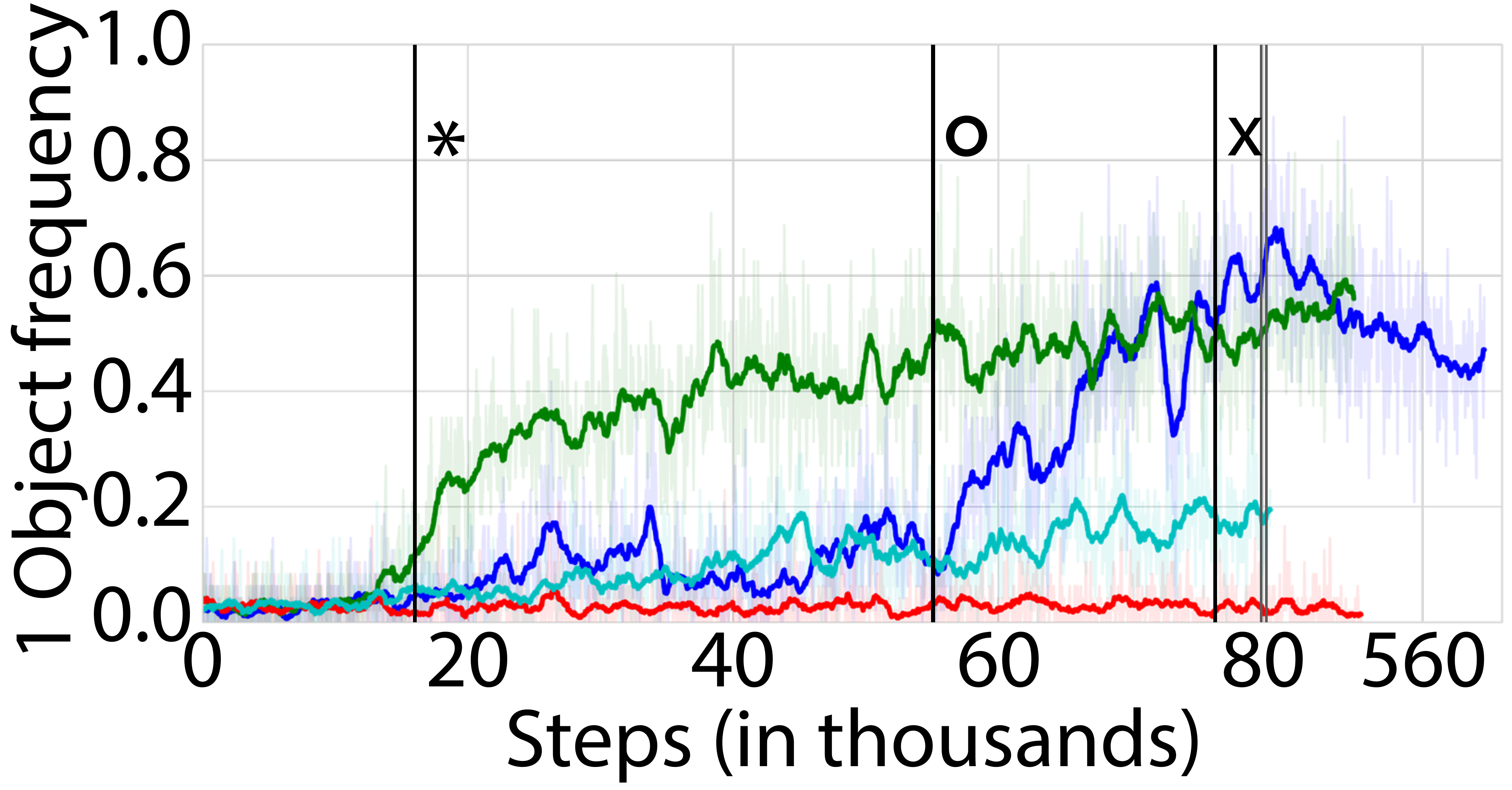}
        &
        \includegraphics[width=0.3\linewidth]{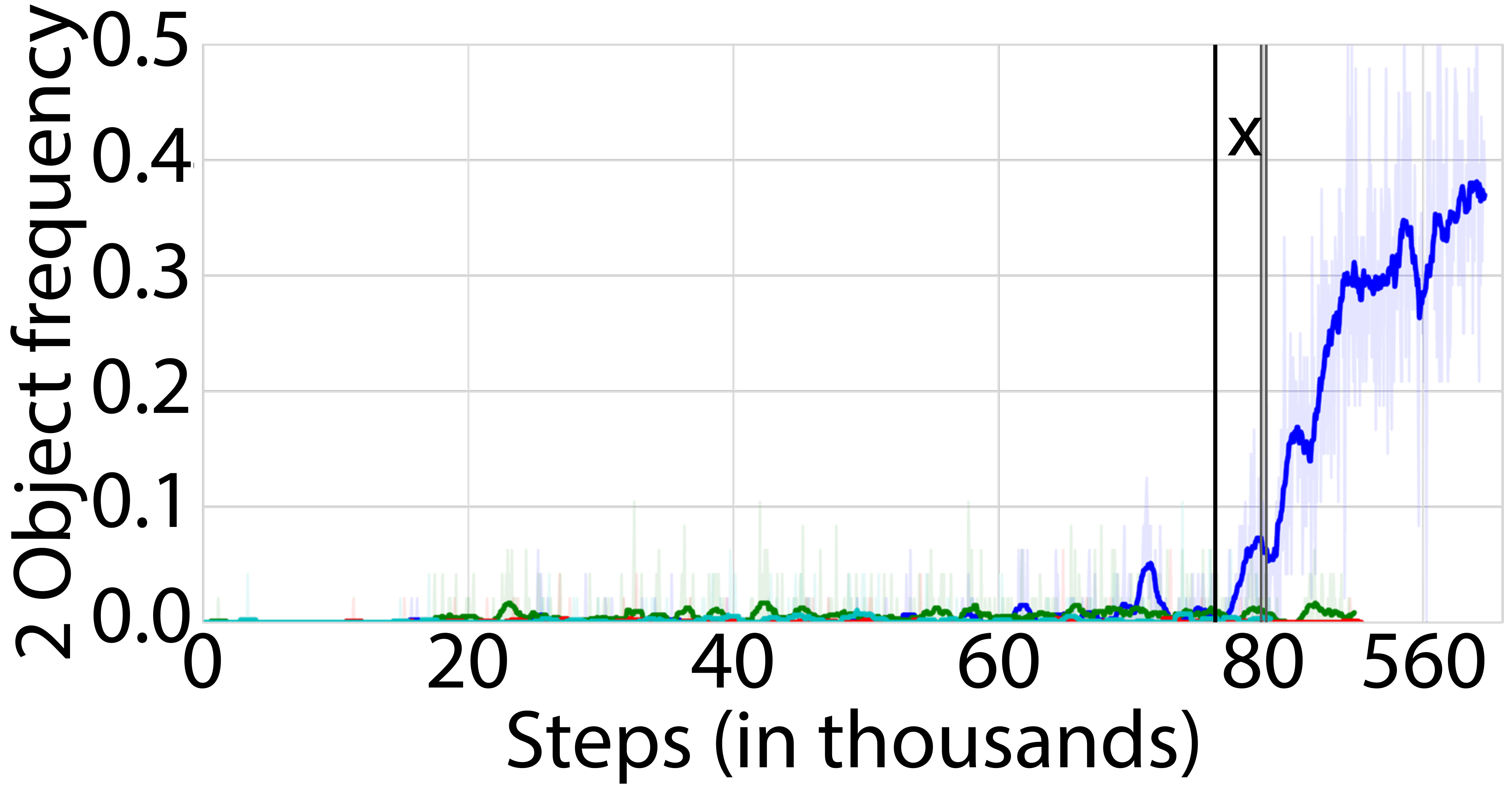} 
        &
        \includegraphics[width=0.3\linewidth]{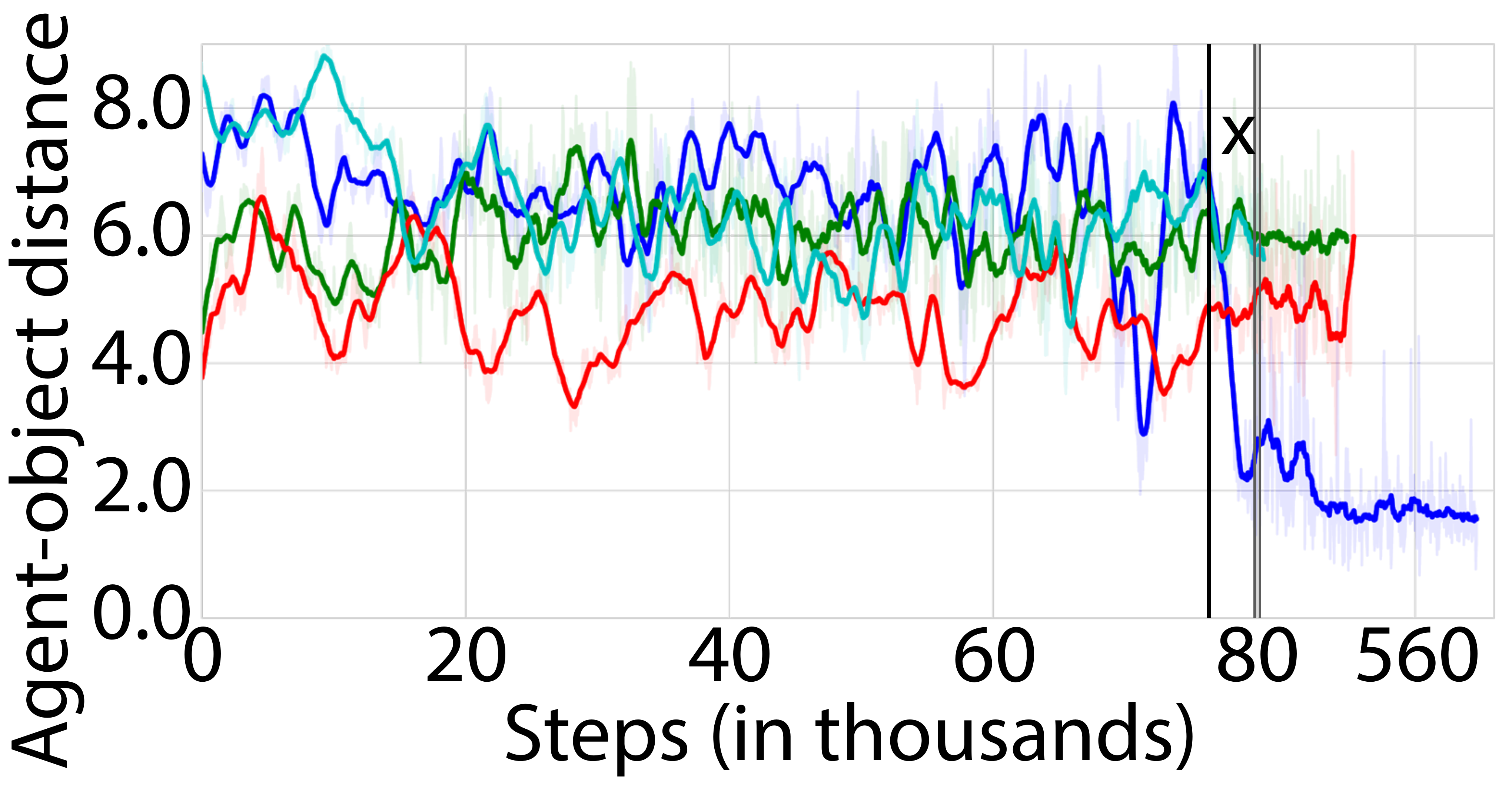}
        \\
        (b) One Object frequency
        &
        (c) Two Object frequency
        &
        (d) Average agent-object distance
        \\
  \end{tabular}
\end{center}
\vspace{-0.3cm}
\caption{\emph{Two-object experiments.} (a) World-model training loss. (b) Percentage of frames in which one object is present. (c) Percentage of frames in which two objects are present. (d) Average distance between agent and objects in Unity units. For this average to be low ($\sim 2.$) both objects must be close to the agent simultaneously. Lighter curves represent unsmoothed batch-mean values.}
\label{fig:2obj_experiments}
\vskip -0.2in
\end{figure*}

\textbf{Multi-object interactions}.
In experiments with multiple objects present, initial learning stages mirror those for the one object experiment (Fig. \ref{fig:2obj_experiments}a) for both ID-SP and LF-SP. The loss temporarily decreases as the agent learns to predict its ego-motion and rises when its attention shifts towards objects, which it then interacts with. However, for ID-SP agents with sufficiently long time horizon (e.g. $T=40$), we robustly observe the emergence of an additional stage in which the loss increases further.  This stage corresponds to the agent gathering and ``playing'' with two objects simultaneously, reflected in a sharp increase in two-object play state frequency (Fig. \ref{fig:2obj_experiments}c), and a decrease in the average distance between the agent and the both objects (Fig. \ref{fig:2obj_experiments}d). We do not observe this additional stage either for ID-SP of shorter time horizon (e.g. $T=5$) or for the LF-SP model even with longer horizons. The ID-SP and LF-SP agents both experience two object play slightly more often than the ID-RP baseline, having achieved substantial one object play time. However, only the ID-SP agent has discovered how to take advantage of the increased difficulty and therefore ``interestingness'' of two object configurations (compare blue with green horizontal line in Fig. \ref{fig:2obj_experiments}a).

\vspace{-.2cm}
\subsection{Dynamics prediction tasks}
\vspace{-.2cm}

We measure the inverse dynamics prediction performance on two held-out validation subsets of data generated from the uncontrolled background distribution of events: (i) an \emph{easy} dataset consisting solely of ego-motion with no play states, and (ii) a \emph{hard} dataset heavily enriched for play states, each with 4800 examples. These data  are collected by executing a random policy in sixteen simulation instances, each containing one object, one for each object type. The hard dataset is the set of examples for which the object is in a play state immediately before the action to be predicted, and the easy dataset is the complement of this.
This measures active learning gains, assessing to what extent the agent self-constructs training data for the hard subset while retaining performance on the easy dataset.

\textbf{Ego-motion learning}.
All aside from random-encoding agents IDRW-RP and IDRW-SP learn ego-motion prediction effectively. 
The ID-RP model quickly converges to a low loss value, where it remains from then on, having effectively learned ego-motion prediction without an antagonistic policy since ego-motion interactions are common in the background random data distribution.
The ID-SP and LF-SP models also learn ego-motion effectively, as seen in the initial decrease of their training losses (Fig.~\ref{fig:1obj_experiments}a) and low loss on the easy ego-motion validation dataset (Fig.~\ref{fig:1obj_experiments}c).

\textbf{Object dynamics prediction}.
Object attention and navigation lead SP agents to substantially different data distributions than baselines.
We evaluate the inverse dynamics prediction performance on the held-out hard object interaction validation set. Here, the ID-SP and LF-SP agents outperform the baselines on predicting the harder object interaction subset by a significant margin, showing that increased object attention translates to improved inverse dynamics prediction (see Fig. \ref{fig:1obj_experiments}d and Table \ref{table:results}).  
Crucially, even though ID-SP and LF-SP have substantially decreased the fraction of time spent on ego-motion interactions (Fig. \ref{fig:1obj_experiments}c), they still retain high performance on that easier sub-task.

\vspace{-.2cm}
\subsection{Task transfers}
\vspace{-.2cm}

We measure the agents' abilities to solve visual tasks for which they were not directly trained, including (i) object presence, (ii) localization, as measured by pixel-wise 2D centroid position, and (iii) 16-way object category recognition. 
We collect data with a random policy from sixteen simulation instances (each with one object, one for each object). For object presence, we subselect examples so as to have an equal number with and without an object in view. 
For localization and category identity (discerning which of the sixteen objects is in view), we take only frames with the object in a play state. These data are split into train (16000 examples), validation (8000 examples), and test (8000 examples) sets. On train sets, we fit elastic net regression/classification for each layer of both world- and self-model encodings, and we use validation sets to select the best-performing model per agent type. These best models are then evaluated on the test sets.
Note that the test sets contain substantial variation in position, pose and size, rendering these tasks nontrivial.
Self-model driven agents substantially outperform alternatives on all three transfer tasks, 
As shown in Table \ref{table:results}, the SP ($T = 5$) agents outperform baselines on inverse dynamics and object presence metrics, while ID-SP outperforms LF-SP on localization and recognition.
Crucially, the ID-RP ablation comparison shows that {\em without an active learning policy, the encoding learned performs comparitively poorly on transfer tasks.}
Interestingly, we find that training with two objects present improves recognition transfer performance as compared to one object scenarios, potentially due to the greater complexity of two-object configurations (Table \ref{table:results}). This is especially notable for the ID-SP ($T = 40$) agent that constructs a substantially increased percentage of two-object events.

\begin{table*}
\caption{
\textbf{Performance comparison.} Ego-motion ($v_{fwd}$, $v_\theta$) and interaction ($f,\tau$) accuracy in \% is compared for play and non-play states. Object frequency, presence and recognition are measured in \% and localization in mean pixel error. Models are trained with one object per room unless stated.}
\vspace{-0.1cm}
\label{table:results}
\vskip 0.15in
\begin{center}
\begin{small}
\begin{sc}
\begin{tabular}{lccccc}
\toprule
Task & IDRW-RP & IDRW-SP & ID-RP & ID-SP & LF-SP \\
\midrule
$v_{fwd}$ accuracy --- Easy  & 65.9 & 56.0 & \textbf{96.0} & 95.3 & 95.3 \\
$v_\theta$ accuracy --- Easy & 82.9 & 75.2 & \textbf{98.7} & 98.4 & 98.5\\
$v_{fwd}$ accuracy --- Hard      & 62.4 & 69.2 & 90.4 & \textbf{95.9} & 95.4 \\
$v_\theta$ accuracy --- Hard    & 79.0 & 80.0 & 95.5 & \textbf{98.2} & \textbf{98.1} \\
$f$ accuracy --- Hard   & 20.8 & 33.1 & 42.1 & \textbf{51.1} & 45.1 \\
$\tau$ accuracy --- Hard   & 20.9 & 32.1 & 41.3 & \textbf{43.2} & \textbf{43.2} \\
Object frequency             & 0.50 & 47.9 & 0.40 & \textbf{61.1} & 12.8 \\
\midrule
Object Presence error               & 4.0 & 3.0 & 0.92 & 0.92 & \textbf{0.60} \\
Localization error [px]   & 15.04 & 10.14 & 5.94 &  \textbf{4.36} &  5.94\\
Recognition accuracy         & 13.0 & 21.99 & 12.3 & \textbf{28.5} & 18.7 \\
\midrule
Recognition acc. -- 2 object training & 12.0 & - & 16.1 & \textbf{39.7} & 21.1 \\
\bottomrule
\end{tabular}
\end{sc}
\end{small}
\end{center}
\vskip -0.4in
\end{table*} 

\section{Discussion}
\vspace{-.2cm}
We have constructed a simple self-supervised mechanism that spontaneously generates a spectrum of emergent naturalistic behaviors via an active learning process, experiencing ``developmental milestones'' of increasing complexity as the agent learns to ``play''. 
The agent first learns the dynamics of its own motion, gets ``bored'', then shifts its attention to locating, moving toward, and interacting with single objects ($*$ in Fig.~\ref{fig:1obj_experiments} and Fig.~\ref{fig:2obj_experiments}).
Once these are better understood ($\circ$ in Fig.~\ref{fig:1obj_experiments} and Fig.~\ref{fig:2obj_experiments}), the agent transitions to gathering multiple objects and learning from their interactions ($\times$ in Fig.~\ref{fig:2obj_experiments}).
This increasingly challenging self-generated curriculum leads to performance gains in the agent's world-model and improved transfer to other useful visual tasks on which the system never received any explicit training signal.
Our ablation studies show that without this active learning policy, world-model accuracy remains poor and visual encodings transfer much less well. 
These results constitute a proof-of-concept that both complex behaviors and useful visual features can arise from simple intrinsic motivations in a three-dimensional physical environment with realistically large and continuous state and action spaces. 

\begin{wrapfigure}{R}{0.75\textwidth}
\vspace{-.4cm}
\begin{center}
\includegraphics[width=.75\textwidth]{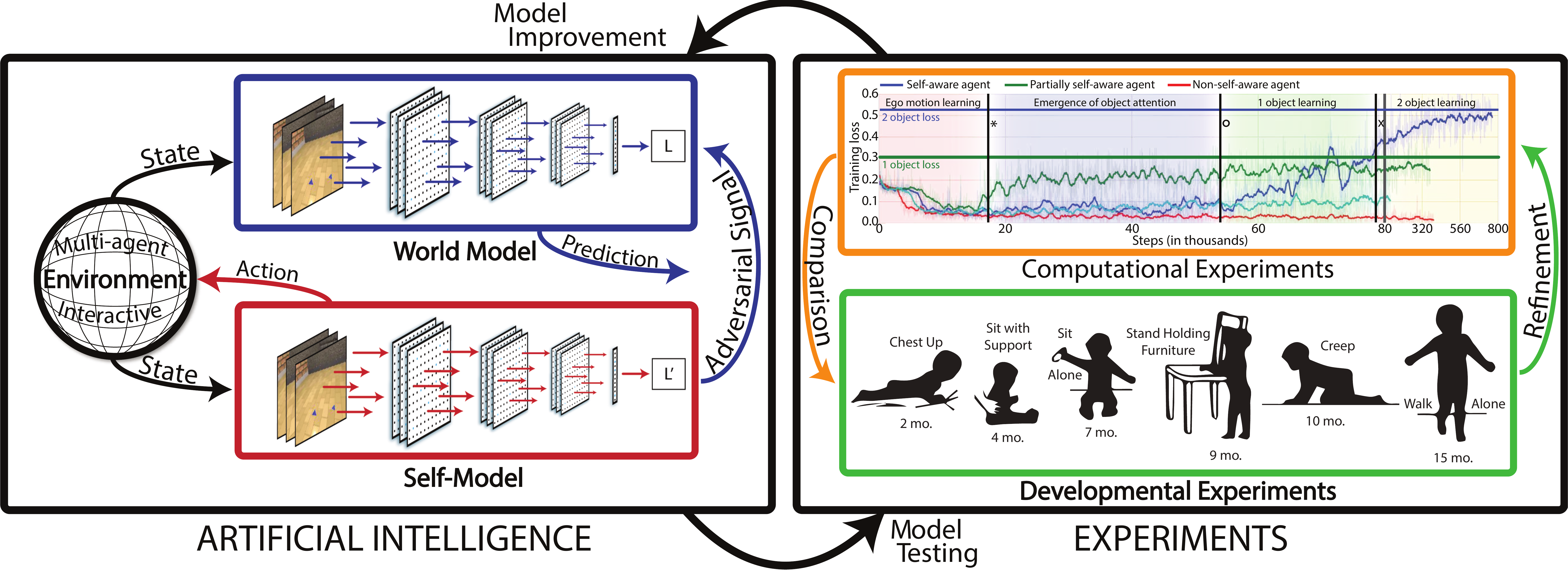}
\end{center}
\vspace{-0.3cm}
\caption{Computational model and human development comparison.}
\label{fig:behavioral_comparison}
\vskip -0.1in
\end{wrapfigure}

In future work, we seek to generate much more sophisticated behaviors than those seen here, including the creation of complex planned trajectories and the building of useful environmental structures. Beyond the objective of building more robustly learning AI, we seek to build computational models that admit precise quantitiative comparisons to the developmental trajectories observed in human children (Figure~\ref{fig:behavioral_comparison}).
To this end, our environment will need better graphics and physics, more varied visual objects, and more realistically embodied robotic agent avatars with articulated actuators and haptic sensors.
From a core algorithms approach, we will need to improve our approach to handling the inherent degeneracy (``white-noise problem'') of our dynamics prediction problems.
The LF-SP agent employs, as discussed in Section~\ref{sec:env_and_architecture}, the technique in \citep{berkeley_mario} aimed at this. 
It was, however, unclear whether this method fully resolved the issue. 
It will likely be necessary to improve both the formulation of the world-model dynamics prediction tasks our agent solves as well as the antagonistic action policies of the agent's self-model.
One approach may be improving our formulation of curiosity from the simple adversarial concept to include additional notions of intrinsic motivation such as learning progress \citep{schmidhuber_formaltheoryoffun,oudeyer2007intrinsic,oudeyer2016evolution}.
More refined future prediction models (e.g.~\citep{mrowca2018flexible}) may also ameliorate degeneracy and lead to more sophisticated behavior.
Finally, including other animate agents in the environment will not only lead to more complex interactions, but potentially also better learning through imitation \citep{ho2016generative}.
In this scenario, the self-model component of our architecture will need to be not only aware of the agent itself, but also make predictions about the actions of other agents --- perhaps providing a connection to the cognitive science of \emph{theory of mind} \citep{saxe2003people}.

\vspace{-.3cm}
\section*{Acknowledgments}
\vspace{-.2cm}
This work was supported by grants from the James S. McDonnell Foundation, Simons Foundation, and Sloan Foundation (DLKY), a Berry Foundation postdoctoral fellowship and Stanford Department of Biomedical Data Science NLM T-15 LM007033-35 (NH), ONR - MURI (Stanford Lead) N00014-16-1-2127 and ONR - MURI (UCLA Lead) 1015 G TA275 (LF).

\bibliography{refs}

\begin{thebibliography}{46}
\providecommand{\natexlab}[1]{#1}
\providecommand{\url}[1]{\texttt{#1}}
\expandafter\ifx\csname urlstyle\endcsname\relax
  \providecommand{\doi}[1]{doi: #1}\else
  \providecommand{\doi}{doi: \begingroup \urlstyle{rm}\Url}\fi

\bibitem[Achiam and Sastry(2017)]{achiam2017surprise}
J.~Achiam and S.~Sastry.
\newblock Surprise-based intrinsic motivation for deep reinforcement learning.
\newblock \emph{CoRR}, abs/1703.01732, 2017.

\bibitem[Agrawal et~al.(2016)Agrawal, Nair, Abbeel, Malik, and
  Levine]{agrawal2016learning}
P.~Agrawal, A.~V. Nair, P.~Abbeel, J.~Malik, and S.~Levine.
\newblock Learning to poke by poking: Experiential learning of intuitive
  physics.
\newblock In \emph{NIPS}, 2016.

\bibitem[Baranes and Oudeyer(2013)]{baranes2013active}
A.~Baranes and P.-Y. Oudeyer.
\newblock Active learning of inverse models with intrinsically motivated goal
  exploration in robots.
\newblock \emph{Robotics and Autonomous Systems}, 61\penalty0 (1):\penalty0
  49--73, 2013.

\bibitem[Begus et~al.(2014)Begus, Gliga, and
  Southgate]{begus_infantslearnwhattheywant}
K.~Begus, T.~Gliga, and V.~Southgate.
\newblock Infants learn what they want to learn: Responding to infant pointing
  leads to superior learning.
\newblock \emph{PLOS ONE}, 9\penalty0 (10):\penalty0 1--4, 10 2014.
\newblock \doi{10.1371/journal.pone.0108817}.

\bibitem[Boeing and Br{\"a}unl(2007)]{boeing2007evaluation}
A.~Boeing and T.~Br{\"a}unl.
\newblock Evaluation of real-time physics simulation systems.
\newblock In \emph{Proceedings of the 5th international conference on Computer
  graphics and interactive techniques in Australia and Southeast Asia}, pages
  281--288. ACM, 2007.

\bibitem[Chentanez et~al.(2005)Chentanez, Barto, and
  Singh]{chentanez2005intrinsically}
N.~Chentanez, A.~G. Barto, and S.~P. Singh.
\newblock Intrinsically motivated reinforcement learning.
\newblock In \emph{NIPS}, pages 1281--1288, 2005.

\bibitem[Ebert et~al.(2017)Ebert, Finn, Lee, and Levine]{ebert2017self}
F.~Ebert, C.~Finn, A.~X. Lee, and S.~Levine.
\newblock Self-supervised visual planning with temporal skip connections.
\newblock In \emph{CoRL}, volume~78, pages 344--356. PMLR, 2017.

\bibitem[Elhamifar et~al.(2013)Elhamifar, Sapiro, Yang, and
  Sastry]{elhamifar2013convex}
E.~Elhamifar, G.~Sapiro, A.~Y. Yang, and S.~S. Sastry.
\newblock A convex optimization framework for active learning.
\newblock In \emph{ICCV}, pages 209--216. IEEE Computer Society, 2013.
\newblock ISBN 978-1-4799-2839-2.

\bibitem[Fantz(1964)]{fantz_visualexperienceininfants}
R.~L. Fantz.
\newblock Visual experience in infants: Decreased attention to familiar
  patterns relative to novel ones.
\newblock \emph{Science}, 146:\penalty0 668--670, 1964.

\bibitem[Finn and Levine(2017)]{finn2017deep}
C.~Finn and S.~Levine.
\newblock Deep visual foresight for planning robot motion.
\newblock In \emph{ICRA}, pages 2786--2793. IEEE, 2017.
\newblock ISBN 978-1-5090-4633-1.

\bibitem[Frank et~al.(2014)Frank, Leitner, Stollenga, Förster, and
  Schmidhuber]{frank2014curiosity}
M.~Frank, J.~Leitner, M.~F. Stollenga, A.~Förster, and J.~Schmidhuber.
\newblock Curiosity driven reinforcement learning for motion planning on
  humanoids.
\newblock \emph{Front. Neurorobot.}, 2014, 2014.

\bibitem[Gilad-Bachrach et~al.(2005)Gilad-Bachrach, Navot, and
  Tishby]{giladbachrach_qbcreal_active}
R.~Gilad-Bachrach, A.~Navot, and N.~Tishby.
\newblock Query by committee made real.
\newblock In \emph{NIPS}, pages 443--450, 2005.

\bibitem[Gopnik et~al.(2009)Gopnik, Meltzoff, and Kuhl]{gopnik_scientistincrib}
A.~Gopnik, A.~Meltzoff, and P.~Kuhl.
\newblock \emph{The Scientist In The Crib: Minds, Brains, And How Children
  Learn}.
\newblock HarperCollins, 2009.
\newblock ISBN 9780061846915.

\bibitem[Gottlieb et~al.(2013)Gottlieb, Oudeyer, Lopes, and
  Baranes]{gottlieb2013information}
J.~Gottlieb, P.-Y. Oudeyer, M.~Lopes, and A.~Baranes.
\newblock Information-seeking, curiosity, and attention: computational and
  neural mechanisms.
\newblock \emph{Trends in cognitive sciences}, 17\penalty0 (11):\penalty0
  585--593, 2013.

\bibitem[Goupil et~al.(2016)Goupil, Romand-Monnier, and
  Kouider]{goupil_infantsask}
L.~Goupil, M.~Romand-Monnier, and S.~Kouider.
\newblock Infants ask for help when they know they don’t know.
\newblock \emph{Proceedings of the National Academy of Sciences}, 113\penalty0
  (13):\penalty0 3492--3496, 2016.
\newblock \doi{10.1073/pnas.1515129113}.

\bibitem[Held et~al.(2017)Held, Geng, Florensa, and Abbeel]{held2017automatic}
D.~Held, X.~Geng, C.~Florensa, and P.~Abbeel.
\newblock Automatic goal generation for reinforcement learning agents.
\newblock \emph{CoRR}, abs/1705.06366, 2017.

\bibitem[Ho and Ermon(2016)]{ho2016generative}
J.~Ho and S.~Ermon.
\newblock Generative adversarial imitation learning.
\newblock In \emph{NIPS}, pages 4565--4573, 2016.

\bibitem[Hong et~al.(2017)Hong, Yeo, Kwak, Lee, and
  Han]{selfsupervision_semanticsegmentation}
S.~Hong, D.~Yeo, S.~Kwak, H.~Lee, and B.~Han.
\newblock Weakly supervised semantic segmentation using web-crawled videos.
\newblock In \emph{CVPR}, pages 2224--2232. IEEE Computer Society, 2017.
\newblock ISBN 978-1-5386-0457-1.

\bibitem[Houthooft et~al.(2016)Houthooft, Chen, Chen, Duan, Schulman, Turck,
  and Abbeel]{houthooft_vime}
R.~Houthooft, X.~Chen, X.~Chen, Y.~Duan, J.~Schulman, F.~D. Turck, and
  P.~Abbeel.
\newblock Vime: Variational information maximizing exploration.
\newblock In \emph{NIPS}, pages 1109--1117, 2016.

\bibitem[Hurley and Oakes(2015)]{hurley2015_petexposure}
K.~B. Hurley and L.~M. Oakes.
\newblock {{E}xperience and distribution of attention: {P}et exposure and
  infants' scanning of animal images}.
\newblock \emph{J Cogn Dev}, 16\penalty0 (1):\penalty0 11--30, Jan 2015.

\bibitem[Hurley et~al.(2010)Hurley, Kovack-Lesh, and
  Oakes]{hurley2010_influenceofpets}
K.~B. Hurley, K.~A. Kovack-Lesh, and L.~M. Oakes.
\newblock {{T}he influence of pets on infants' processing of cat and dog
  images}.
\newblock \emph{Infant Behav Dev}, 33\penalty0 (4):\penalty0 619--628, Dec
  2010.

\bibitem[Jaderberg et~al.(2016)Jaderberg, Mnih, Czarnecki, Schaul, Leibo,
  Silver, and Kavukcuoglu]{jaderberg2016reinforcement}
M.~Jaderberg, V.~Mnih, W.~M. Czarnecki, T.~Schaul, J.~Z. Leibo, D.~Silver, and
  K.~Kavukcuoglu.
\newblock Reinforcement learning with unsupervised auxiliary tasks.
\newblock \emph{CoRR}, abs/1611.05397, 2016.

\bibitem[Kalchbrenner et~al.(2017)Kalchbrenner, van~den Oord, Simonyan,
  Danihelka, Vinyals, Graves, and Kavukcuoglu]{kalchbrenner2016video}
N.~Kalchbrenner, A.~van~den Oord, K.~Simonyan, I.~Danihelka, O.~Vinyals,
  A.~Graves, and K.~Kavukcuoglu.
\newblock Video pixel networks.
\newblock In \emph{ICML}, volume~70 of \emph{JMLR Workshop and Conference
  Proceedings}, pages 1771--1779. JMLR.org, 2017.

\bibitem[Kidd et~al.(2012)Kidd, Piantadosi, and Aslin]{kidd2012_goldilocks}
C.~Kidd, S.~T. Piantadosi, and R.~N. Aslin.
\newblock The goldilocks effect: Human infants allocate attention to visual
  sequences that are neither too simple nor too complex.
\newblock \emph{PLOS ONE}, 7\penalty0 (5):\penalty0 1--8, 05 2012.
\newblock \doi{10.1371/journal.pone.0036399}.

\bibitem[Kingma and Ba(2014)]{kingma2014adam}
D.~P. Kingma and J.~Ba.
\newblock Adam: A method for stochastic optimization.
\newblock \emph{CoRR}, abs/1412.6980, 2014.

\bibitem[Kulkarni et~al.(2016)Kulkarni, Narasimhan, Saeedi, and
  Tenenbaum]{kulkarni_hierarchical}
T.~D. Kulkarni, K.~Narasimhan, A.~Saeedi, and J.~Tenenbaum.
\newblock Hierarchical deep reinforcement learning: Integrating temporal
  abstraction and intrinsic motivation.
\newblock In \emph{NIPS}, pages 3675--3683, 2016.

\bibitem[Lin(1992)]{Lin_rlmethods_experiencereplay}
L.-J. Lin.
\newblock \emph{Reinforcement Learning for Robots Using Neural Networks}.
\newblock PhD thesis, Pittsburgh, PA, USA, 1992.
\newblock UMI Order No. GAX93-22750.

\bibitem[Machado et~al.(2017)Machado, Bellemare, and
  Bowling]{machado2017laplacian}
M.~C. Machado, M.~G. Bellemare, and M.~Bowling.
\newblock A laplacian framework for option discovery in reinforcement learning.
\newblock \emph{arXiv preprint arXiv:1703.00956}, 2017.

\bibitem[Mitash et~al.(2017)Mitash, Bekris, and
  Boularias]{selfsupervision_pose}
C.~Mitash, K.~E. Bekris, and A.~Boularias.
\newblock A self-supervised learning system for object detection using physics
  simulation and multi-view pose estimation.
\newblock In \emph{IROS}, pages 545--551. IEEE, 2017.
\newblock ISBN 978-1-5386-2682-5.

\bibitem[Mnih et~al.(2016)Mnih, Badia, Mirza, Graves, Harley, Lillicrap,
  Silver, and Kavukcuoglu]{mnih_asynchronousrl}
V.~Mnih, A.~P. Badia, M.~Mirza, A.~Graves, T.~Harley, T.~P. Lillicrap,
  D.~Silver, and K.~Kavukcuoglu.
\newblock Asynchronous methods for deep reinforcement learning.
\newblock In \emph{ICML}, ICML'16, pages 1928--1937. JMLR.org, 2016.

\bibitem[Mrowca et~al.(2018)Mrowca, Zhuang, Wang, Haber, Fei-Fei, Tenenbaum,
  and Yamins]{mrowca2018flexible}
D.~Mrowca, C.~Zhuang, E.~Wang, N.~Haber, L.~Fei-Fei, J.~B. Tenenbaum, and D.~L.
  Yamins.
\newblock Flexible neural representation for physics prediction.
\newblock \emph{arXiv preprint arXiv:1806.08047}, 2018.

\bibitem[Noroozi and Favaro(2016)]{jigsaw_puzzles}
M.~Noroozi and P.~Favaro.
\newblock Unsupervised learning of visual representations by solving jigsaw
  puzzles.
\newblock In \emph{ECCV (6)}, volume 9910 of \emph{Lecture Notes in Computer
  Science}, pages 69--84. Springer, 2016.
\newblock ISBN 978-3-319-46465-7.

\bibitem[Oudeyer and Smith(2016)]{oudeyer2016evolution}
P.-Y. Oudeyer and L.~B. Smith.
\newblock How evolution may work through curiosity-driven developmental
  process.
\newblock \emph{Topics in Cognitive Science}, 8\penalty0 (2):\penalty0
  492--502, 2016.

\bibitem[Oudeyer et~al.(2007)Oudeyer, Kaplan, and Hafner]{oudeyer2007intrinsic}
P.-Y. Oudeyer, F.~Kaplan, and V.~V. Hafner.
\newblock Intrinsic motivation systems for autonomous mental development.
\newblock \emph{IEEE transactions on evolutionary computation}, 11\penalty0
  (2):\penalty0 265--286, 2007.

\bibitem[Pathak et~al.(2017)Pathak, Agrawal, Efros, and
  Darrell]{berkeley_mario}
D.~Pathak, P.~Agrawal, A.~A. Efros, and T.~Darrell.
\newblock Curiosity-driven exploration by self-supervised prediction.
\newblock In \emph{ICML}, volume~70 of \emph{JMLR Workshop and Conference
  Proceedings}, pages 2778--2787. JMLR.org, 2017.

\bibitem[Popov et~al.(2017)Popov, Heess, Lillicrap, Hafner, Barth-Maron,
  Vecerik, Lampe, Tassa, Erez, and Riedmiller]{popov2017data}
I.~Popov, N.~Heess, T.~Lillicrap, R.~Hafner, G.~Barth-Maron, M.~Vecerik,
  T.~Lampe, Y.~Tassa, T.~Erez, and M.~Riedmiller.
\newblock Data-efficient deep reinforcement learning for dexterous
  manipulation.
\newblock \emph{arXiv preprint arXiv:1704.03073}, 2017.

\bibitem[Saxe and Kanwisher(2003)]{saxe2003people}
R.~Saxe and N.~Kanwisher.
\newblock People thinking about thinking people: the role of the
  temporo-parietal junction in “theory of mind”.
\newblock \emph{Neuroimage}, 19\penalty0 (4):\penalty0 1835--1842, 2003.

\bibitem[Schmidhuber(2010)]{schmidhuber_formaltheoryoffun}
J.~Schmidhuber.
\newblock Formal theory of creativity, fun, and intrinsic motivation
  (1990-2010).
\newblock \emph{IEEE Trans. Autonomous Mental Development}, 2\penalty0
  (3):\penalty0 230--247, 2010.

\bibitem[Sener and Savarese(2017)]{sener2017active}
O.~Sener and S.~Savarese.
\newblock A geometric approach to active learning for convolutional neural
  networks.
\newblock \emph{CoRR}, abs/1708.00489, 2017.

\bibitem[Settles(2011)]{settles2011_active}
B.~Settles.
\newblock \emph{Active Learning}, volume~18.
\newblock Morgan \& Claypool Publishers, 2011.

\bibitem[Singh et~al.(2010)Singh, Lewis, Barto, and
  Sorg]{singh2010intrinsically}
S.~P. Singh, R.~L. Lewis, A.~G. Barto, and J.~Sorg.
\newblock Intrinsically motivated reinforcement learning: An evolutionary
  perspective.
\newblock \emph{IEEE Trans. Autonomous Mental Development}, 2\penalty0
  (2):\penalty0 70--82, 2010.

\bibitem[Sokolov(1963)]{sokolov1963perception}
E.~Sokolov.
\newblock \emph{Perception and the conditioned reflex}.
\newblock Pergamon Press, 1963.

\bibitem[Spyros~Gidaris(2018)]{rotnet}
N.~K. Spyros~Gidaris, Praveer~Singh.
\newblock Unsupervised representation learning by predicting image rotations.
\newblock \emph{ICLR}, 2018.

\bibitem[Twomey and Westermann(2017)]{twomey_curiositybasedlearning}
K.~E. Twomey and G.~Westermann.
\newblock {{C}uriosity-based learning in infants: a neurocomputational
  approach}.
\newblock \emph{Dev Sci}, Oct 2017.

\bibitem[Wang et~al.(2017)Wang, Zhang, Li, Zhang, and Lin]{wang2016cost}
K.~Wang, D.~Zhang, Y.~Li, R.~Zhang, and L.~Lin.
\newblock Cost-effective active learning for deep image classification.
\newblock \emph{IEEE Trans. Circuits Syst. Video Techn.}, 27\penalty0
  (12):\penalty0 2591--2600, 2017.

\bibitem[Zhang et~al.(2016)Zhang, Isola, and Efros]{colorization}
R.~Zhang, P.~Isola, and A.~A. Efros.
\newblock Colorful image colorization.
\newblock In B.~Leibe, J.~Matas, N.~Sebe, and M.~Welling, editors, \emph{ECCV
  (3)}, volume 9907 of \emph{Lecture Notes in Computer Science}, pages
  649--666. Springer, 2016.
\newblock ISBN 978-3-319-46486-2.

\end{thebibliography}
\bibliographystyle{abbrvnat}

\cleardoublepage

\section{Supplementary}

In Section~\ref{sec:train_details}, we give explicit training details. In Section~\ref{sec:model_details}, we give explicit architecture details and describe the losses applied to each component. In Section~\ref{sec:object_details}, we describe and depict the objects used in the environment. In Section~\ref{sec:env_stability}, we give the results of an experiment in which we halve the room's dimensions and the agents' maximum speeds, demonstrating the stability of our main results under these changes. Finally, in Section~\ref{sec:object_frequencies}, we examine the frequencies with which the agent interacts with each distinct object.

\subsection{Training details} \label{sec:train_details}

Our training procedure incorporates asynchronous methods \citep{mnih_asynchronousrl} and experience replay \citep{Lin_rlmethods_experiencereplay} with a small buffer, with data gathering threads accumulating histories of data and update threads computing gradients off of shuffled data from a buffer. We instantiate a world-model $\omega_{\theta}$ and self-model $\Lambda_{\phi}$, each with Xavier initialization. The architecture is used to collect data and world-model loss results with $N_e$ environments $\operatorname{env}_{k}$ in parallel. A separate thread performs updates using data from its $N_e$ environments, syncing with the global weights, computing gradients, and updating weights with gradients.

Each data collection thread takes $\operatorname{gather\_per\_batch}$ steps in between enqueueing a batch of size $\operatorname{batch\_size} / N_e$. Each scene lasts a number of environment steps chosen uniformly at random within $[\operatorname{scene\_length\_l\_bound}, \operatorname{scene\_length\_u\_bound})$. At the beginning of each scene, objects are randomly chosen from our sixteen pairs, and the agent and object(s) are placed at positions uniformly at random in the room of size 10x10 Unity units (units defined by the Unity development platform, above referred to as ``meters''). The objects are placed at random orientations just above the ground and fall at the beginning of the scene. The agent is placed upright looking in a random direction. Each maintains an history buffer $h_{k}$ upon which it stores observations (obtained from the environment), actions (chosen by the policy), and world-model loss (computed on data as soon as it is gathered). Policies are computed by sampling $K$ actions uniformly at random, obtaining the policy $\pi$ probabilities on each sample as described in Section~\ref{sec:env_and_architecture}, and sampling from this $K$-way discrete distribution. Batches are constructed from slices of data in the history buffer starting at uniformly randomly-chosen times and are placed in FIFO Queues $Q_{k}$.

The update thread concatenates batches dequeued from $Q_{k}$ for $k = 1 \ldots N_e$, and computes losses and gradients. Note that, as outlined in Section~\ref{sec:env_and_architecture}, different variables have different corresponding losses. For example, in the LF case, there is an auxiliary ID prediction task with variables $\theta_{ID}$ that receive gradient updates from $L_{ID}$, separately from the LF prediction task with variables $\theta_{LF}$ that receive gradient updates from $L_{LF}$. In either the ID or LF cases, the self-model has variables $\psi$ that receive gradient updates from $L_{\Lambda}$ which computes true-values from world-model losses stored in the data collection loop. Gradients are applied to the weights using an Adam optimizer with given $\operatorname{learning\_rate}$.

Except where explicitly specified, we take $N_e = 16$, $\operatorname{initial\_gather} = 250$, $\operatorname{batch\_size} = 32$, $\operatorname{gather\_per\_batch} = 3$ (so with 16 environments, 48 steps are taken in between each batch update), $K = 1000$, and $\operatorname{learning\_rate} = .0001$. Despite self-model true values depending on the policy the agent chooses, we find training to be stable for small experience replay buffers of around $100-1000$ environment steps.

\begin{algorithm}

\SetKwInOut{Init}{Init}

\Init{\\
Dynamics prediction problem $\operatorname{D}, \operatorname{In}, \operatorname{Out}, \iota: \operatorname{D} \rightarrow \operatorname{In}, \tau: \operatorname{D} \rightarrow \operatorname{Out}, L$\\
World-model $\omega_{\theta}$\\
Self-model $\Lambda_{\phi}$\\
Environments $\operatorname{env}_{k}$ for $k = 1 \ldots N_e$\\
batch FIFO Queues $Q_{k}(\mathrm{capacity} = c)$ for $k = 1 \ldots N_e$\\
History lists $h_{k}(\mathrm{length} = \operatorname{initial\_gather})$ for $k = 1 \ldots N_e$\\
gather\_per\_batch, scene\_length\_l\_bound, scene\_length\_u\_bound, batch\_size\\
action\_dim (8 in 1-object case, 14 in 2-object case)\\
number of actions to sample $K$\\
summary map $\sigma$\\
learning\_rate
}

\end{algorithm}

\begin{algorithm}
\underline{Run gather threads for each $\operatorname{env}_{k}$, in parallel.}\\
\Begin{
Fill history list $h_{k}$ with null observations, actions, and losses.
  
  \While{True}{
    $\operatorname{num\_this\_batch} = 0$\\
    \While{$\operatorname{num\_this\_batch} < \operatorname{gather\_per\_batch}$
     or $\operatorname{total\_gathered} < \operatorname{history\_len}$}{
      
      \underline{reset scene if needed}\\
      \Indp\If{$\operatorname{num\_this\_scene} \geq \operatorname{scene\_length}$}{
        $\operatorname{observation} = \operatorname{env}_k$.set\_new\_scene()\\
        delete oldest from history $h_{k}$\\
        store observation, null action, and zero loss in $h_{k}$\\
        $\operatorname{num\_this\_scene} = 0$\\
        $\operatorname{scene\_length} \sim \operatorname{Uniform}(\operatorname{scene\_length\_l\_bound}, \operatorname{scene\_length\_u\_bound})$
      }
      \Indm\underline{take an action}\\
      \Indp\Begin{
        $\operatorname{action\_sample} \sim \operatorname{Uniform}([-1, 1]^{K \times \operatorname{action\_dim}})$\\
        \For{$i = 0 \ldots K - 1$}{
          $(p_1, p_2 \ldots p_T)[i] = \Lambda_{\phi}(\operatorname{action\_sample}[i], \mathrm{last\ two\ observations})$, \\
        }
        policy $\pi(i | \mathrm{current\ state}) = \exp(\beta \sigma((p_1, p_2 \ldots p_T)[i])),$ normalized over $i$\\
        sample $\operatorname{i\_chosen} \sim \pi(i | \mathrm{current\ state})$\\
        $\operatorname{action\_chosen} = \operatorname{action\_sample}[i]$\\
        $\operatorname{observation} = \operatorname{env}_k$.step(action\_chosen)
      }

      \Indm\underline{calculate $\omega_{\theta}$ loss}\\
      \Indp$\operatorname{world-model\ prediction} = \omega_{\theta}(\operatorname{\iota}(\mathrm{most\ recent\ history\ slice}))$\\
      $\operatorname{world-model\ loss} = L(\operatorname{world-model\ prediction}, \operatorname{\tau}(\mathrm{most\ recent\ history\ slice}))$
      
      \Indm\underline{manage history}\\
      \Indp delete oldest from history $h_{k}$\\
      store observation, action\_chosen, world-model loss\\

      \Indm$\mathrm{num\_this\_batch} \leftarrow \mathrm{num\_this\_batch} + 1$\\
      $\mathrm{total\_gathered} \leftarrow \mathrm{total\_gathered} + 1$\\
      $\mathrm{num\_this\_scene} \leftarrow \mathrm{num\_this\_scene} + 1$\\
    }

    \underline{store batch for update}\\
    \Indp choose $\mathrm{batch\_size} / N_e$ slices of $h_{k}$\\
    $\mathrm{batch} = \operatorname{\iota}(\text{slices}), \operatorname{\tau}(\text{slices}), \operatorname{world-model\ losses}$\\
    $Q.\mathrm{enqueue}(\mathrm{batch})$
  }

}
\end{algorithm}

\begin{algorithm}
\underline{Run update thread in parallel with gather threads.}\\
\While{True}{
\For{$k = 1 \ldots N_e$}{
  $\mathrm{batch\_k} = Q_{k}.\mathrm{dequeue}()$
}
$\mathrm{batch} = \mathrm{concatenate}(\mathrm{batch\_k\ for\ }k = 1 \ldots N_e)$\\
compute loss(es) and gradient for $\omega_{\theta}$ (including auxiliary ID model for LF task)\\
compute loss and gradient for $\Lambda_{\phi}$ using cached losses in batch\\
update $\theta$ and $\phi$ with computed gradients using Adam(learning\_rate)\\
}
\end{algorithm}

\subsection{Model architectures and losses} \label{sec:model_details}

\begin{figure*}
\begin{center}
\includegraphics[width=\textwidth]{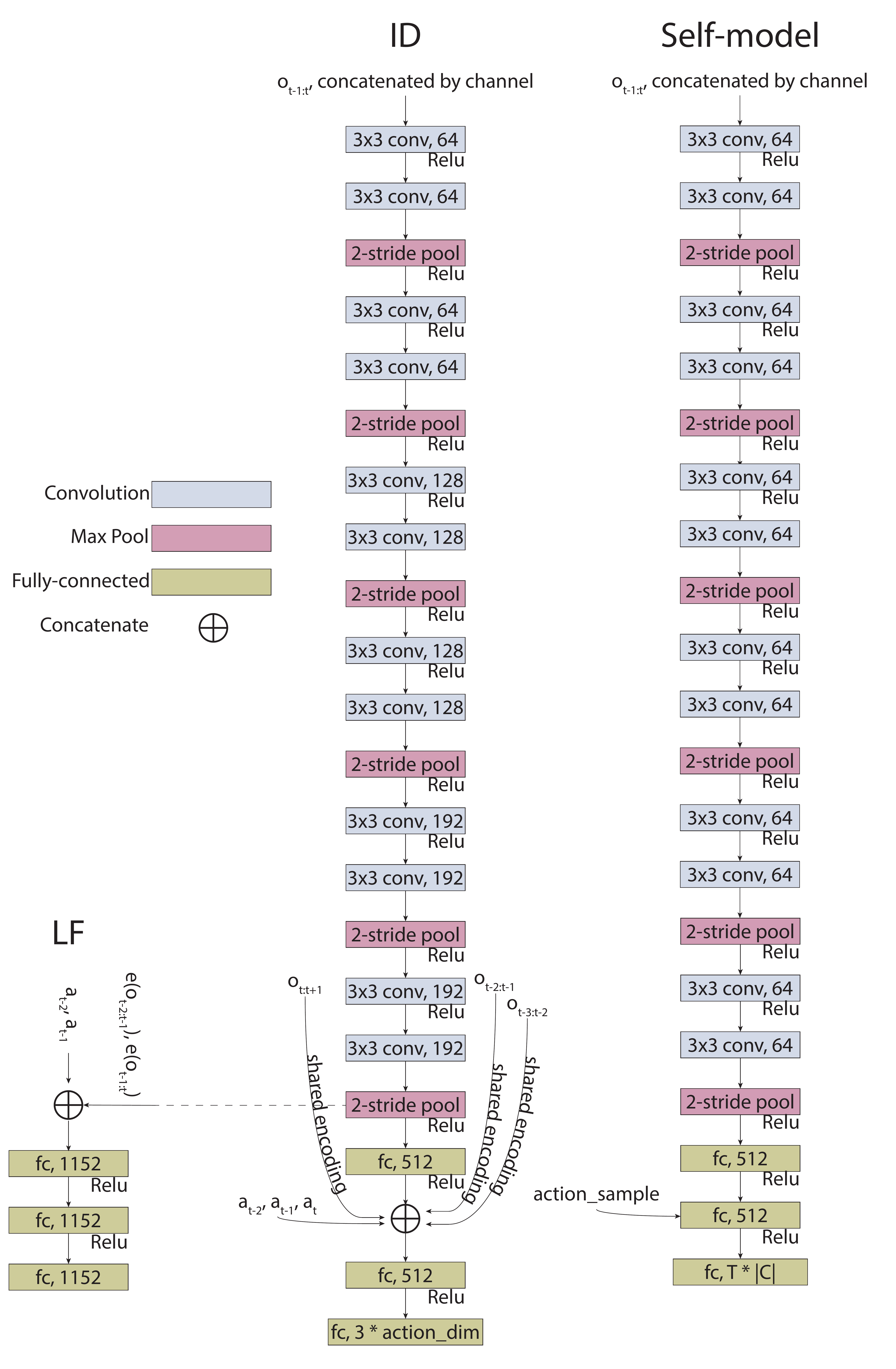}
\end{center}
\vspace{-0.3cm}
\caption{The ID, LF, and self-model architectures.}
\label{fig:explicit_architecture}
\vskip -0.1in
\end{figure*}

We use convolutional neural networks as the base architecture to learn both world-models $\omega_\theta$ and self-models $\Lambda_\psi$.
In our experiments, these networks have an encoding structure with a common architecture involving twelve convolutional layers, two-stride max pools every other layer, and one fully-connected layer, to encode all states into a lower-dimensional latent space, with shared weights across time.
For the inverse dynamics task, the top encoding layer of the network is combined with actions $\{a_{t'}\ |\  t' \neq t\}$, fed into a two-layer fully-connected network, on top of which a softmax classifier is used to predict action $a_t$. 
For the latent space future prediction task, the top convolutional layer of $\omega^{ID}_{\theta_{ID}}$ is used as the latent space $\mathcal{L}$, and the latent model $\omega^{LF}_{\theta_{LF}}$ is parametrized by a fully-connected network that receives, in addition to past encoded images, past actions. See Figure~\ref{fig:explicit_architecture} for a graphical representation.

The ID model (whether or not it is auxiliary to the LF world-model) is supervised by loss $L_{ID}$ in which we make a 3-class classification task by thresholding each dimension of the action by $-.1$ and $.1$:
\begin{equation*}
\operatorname{thresh}(a)_i = 1_{a_i > -.1} + 1_{a_i > .1}, i = 1 \ldots \mathrm{action\_dim}
\end{equation*}
and then averaging softmax cross-entropy loss over each dimension.
The LF model, if used, is supervised by $\ell_2$ loss.

The self-model is supervised by thresholding world-model losses computed in the data gathering loop (Section~\ref{sec:train_details}) by $c \in C$:
\begin{equation*}
\operatorname{thresh}_C(l_t) = \sum_{c \in C} 1_{l > c}
\end{equation*}
and averaging softmax cross-entropy loss over $T$ successive timesteps. 
In the 1-object setting, we took $C = \{.28\}$ for the ID-SP case and $C = \{.13\}$ for the LF-SP case, tuned for object attention (in practice, this appears to matter only in that it satisfies a constraint: below almost all 1-object play losses, for a ID-RP/LF-RP model, but above almost all ego-motion losses, between the first 10000-30000 steps). In the 2-object setting, we chose $C = \{.28, .44, .59\}$ for ID-SP and $C = \{.13, .26, .59\}$ for LF-SP.

To summarize the optimization criteria, in the ID-only case (ID-SP), two objectives are optimized:
\begin{equation*}
\min_{\theta_{ID}} L_{ID} + \min_{\psi} L_{\Lambda, ID},
\end{equation*}
where $L_{ID}$ is a sum, across dimension, of softmax cross-entropy losses on 3-way discretizations of each action dimension, and $L_{\Lambda, ID}$ is a sum, across $T$ timesteps, of softmax cross-entropy losses on $C$-way discretizations of $\omega^{ID}$ loss.
In the LF case (LF-SP), three objectives are optimized:
\begin{equation*}
\min_{\theta_{ID}} L_{ID} + \min_{\theta_{LF}} L_{LF} + \min_{\psi} L_{\Lambda, LF},
\end{equation*}
where $L_{LF}$ is $\ell_2$-loss on the latent space and $L_{\Lambda, LF}$ is like $L_{\Lambda, ID}$ but with $\omega^{ID}$ loss replaced with $\omega^{LF}$.

\subsection{Object details} \label{sec:object_details}

In Figure~\ref{fig:objectablation}, we depict the objects used and give a breakdown of play frequency per object. Shapes are given equal mass (1 Unity unit of mass) and are blue of the same texture. Their geometries consists of varied aspect ratios of four types of shape: sphere, cube, cone, and cylinder, with four per type. 

\begin{figure}
\begin{center}
  \begin{tabular}[c]{cccc}
    \includegraphics[width=0.23\textwidth,height=.23\textwidth, keepaspectratio]{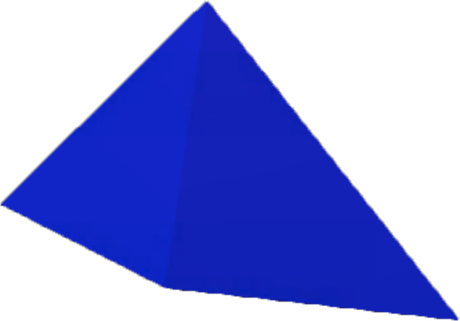}
    &
    \includegraphics[width=0.23\textwidth,height=.23\textwidth, keepaspectratio]{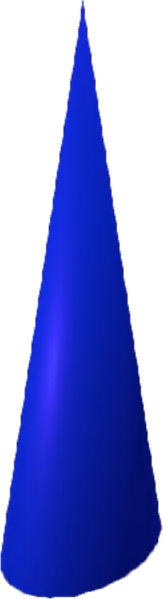}
    &
    \includegraphics[width=0.23\textwidth,height=.23\textwidth, keepaspectratio]{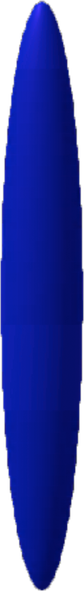}
    &
    \includegraphics[width=0.23\textwidth,height=.23\textwidth, keepaspectratio]{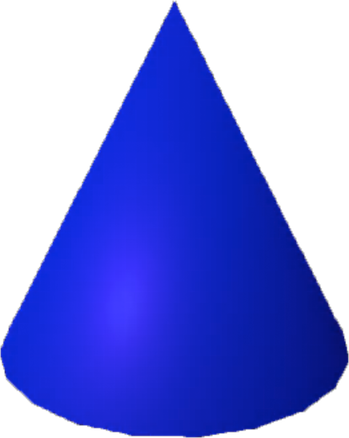}
    \\
    Arrow & Asymmetric cone & Capsule & Cone \\
    \includegraphics[width=0.23\textwidth,height=.23\textwidth, keepaspectratio]{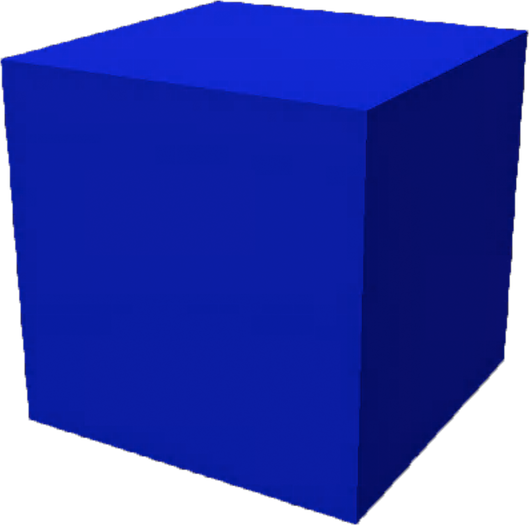}
    &
    \includegraphics[width=0.23\textwidth,height=.23\textwidth, keepaspectratio]{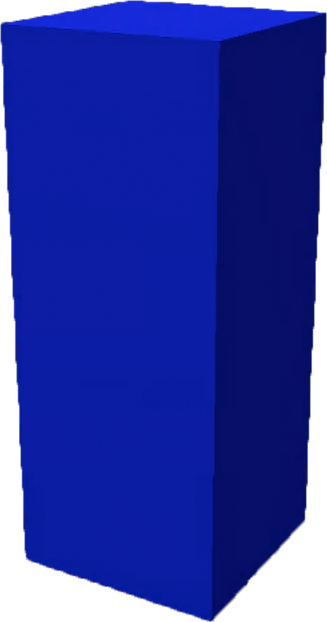}
    &
    \includegraphics[width=0.23\textwidth,height=.23\textwidth, keepaspectratio]{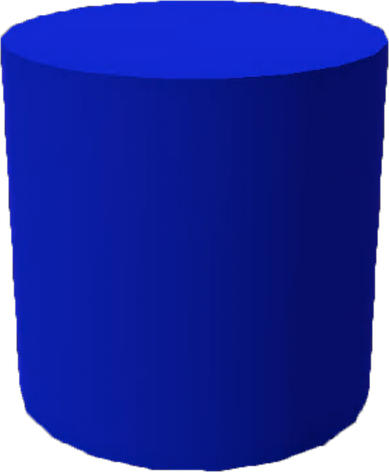}
    &
    \includegraphics[width=0.23\textwidth,height=.23\textwidth, keepaspectratio]{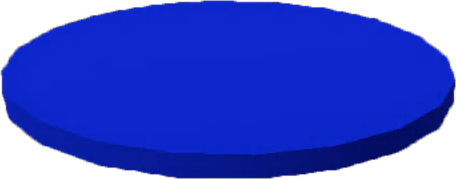}
    \\
    Cube & Cuboid & Cylinder & Disk \\
    \includegraphics[width=0.23\textwidth,height=.23\textwidth, keepaspectratio]{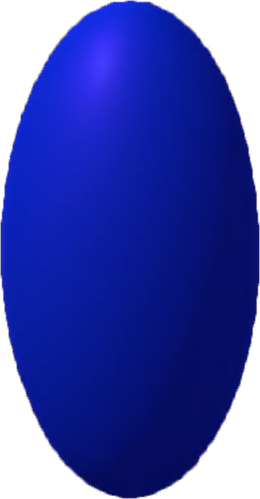}
    &
    \includegraphics[width=0.23\textwidth,height=.23\textwidth, keepaspectratio]{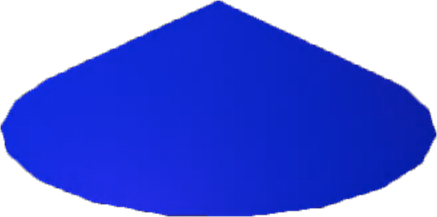}
    &
    \includegraphics[width=0.23\textwidth,height=.23\textwidth, keepaspectratio]{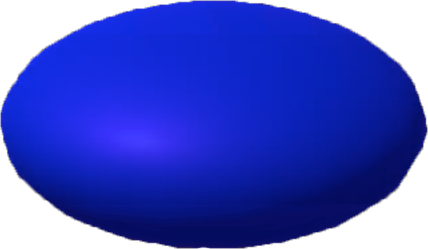}
    &
    \includegraphics[width=0.23\textwidth,height=.23\textwidth, keepaspectratio]{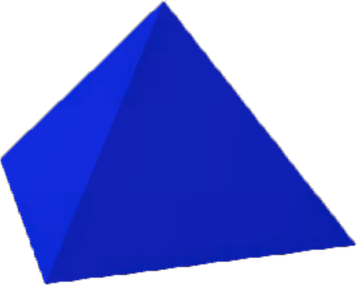}
    \\
    Ellipsoid & Flat cone & Mentos & Pyramid \\ 
    \includegraphics[width=0.23\textwidth,height=.23\textwidth, keepaspectratio]{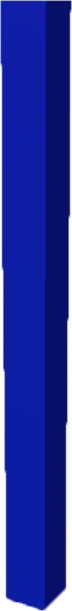}
    &
    \includegraphics[width=0.23\textwidth,height=.23\textwidth, keepaspectratio]{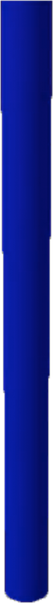}
    &
    \includegraphics[width=0.23\textwidth,height=.23\textwidth, keepaspectratio]{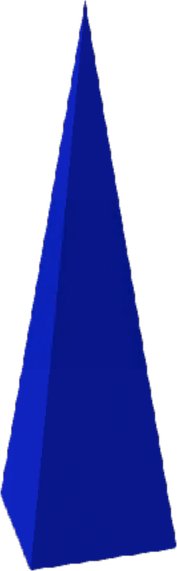}
    &
    \includegraphics[width=0.23\textwidth,height=.23\textwidth, keepaspectratio]{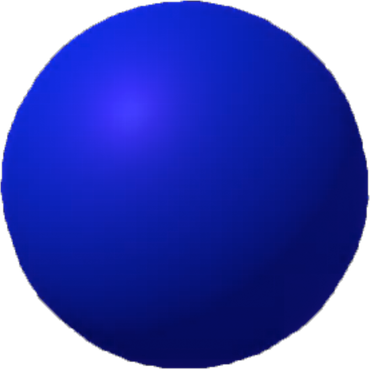}
    \\
    Rectangular stick & Round stick & Spear & Sphere \\
   \end{tabular}
\end{center}
\label{fig:objectablation}
\caption{The objects.}
\end{figure}

\subsection{Stability under varied setups.} \label{sec:env_stability}

\begin{figure}
\begin{center}
\includegraphics[width=\textwidth]{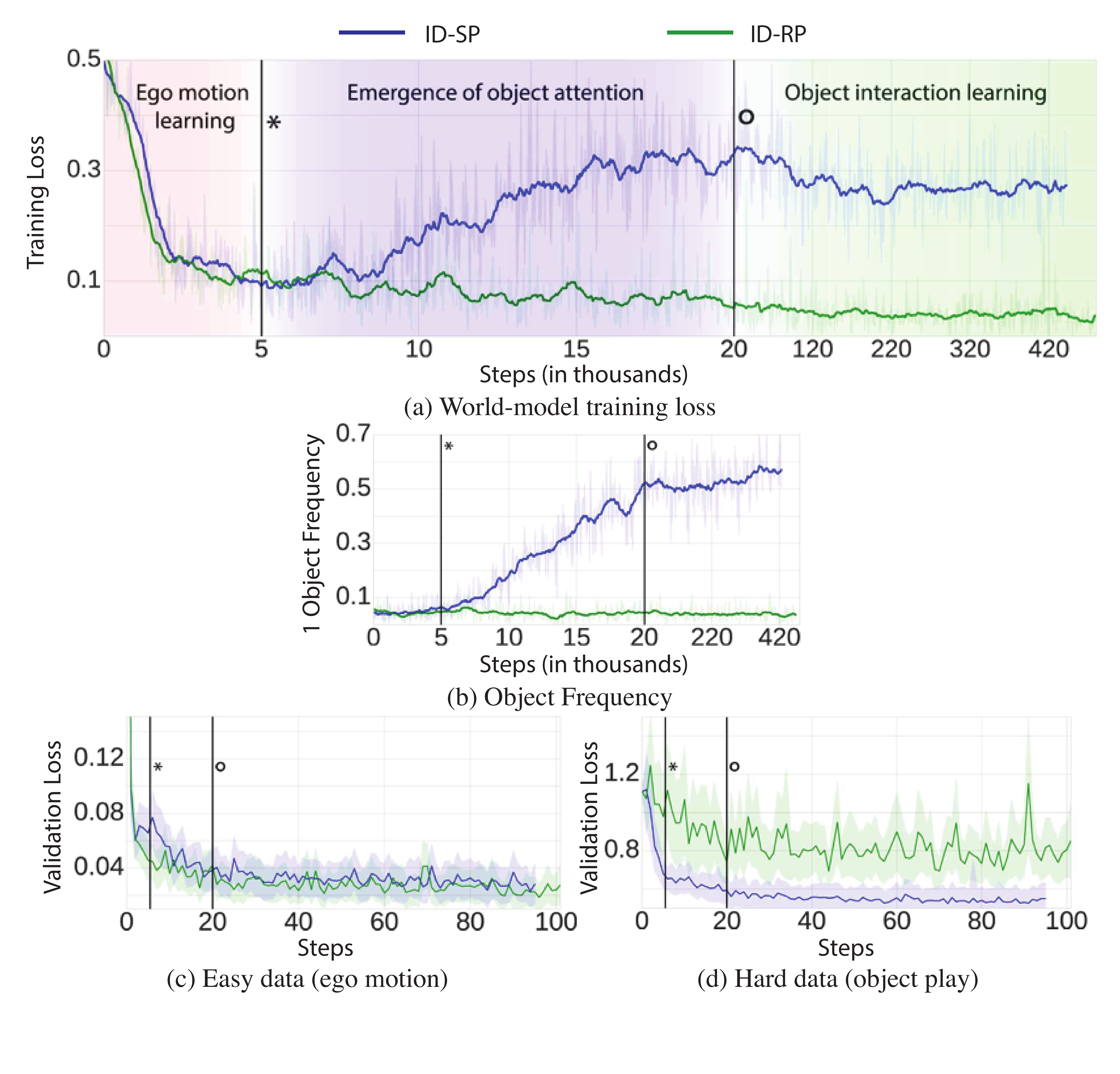}
\end{center}
\vspace{-0.3cm}
\caption{\textbf{Single-object experiments, smaller room and speed.} (a) World-model training loss. (b) Percentage of frames in which an object is present. (c) World-model test-set loss on ``easy'' ego-motion-only data, with no objects present. (d) World-model test-set loss on ``hard'' validation data, with object present, where agent must solve object physics prediction. This experiment differs from those in the main text (compare with Figure~\ref{fig:1obj_experiments}) by halving both the room size and maximum ego-motion speeds while keeping the maximum interaction distance fixed.}
\label{fig:small_room}
\vskip -0.1in
\end{figure}

In this section, we present results in which we vary both the environment and the agent's maximum ego-motions, demonstrating stability under this change of emergent ``developmental milestones'' and dynamics prediction problem performance gains under the antagonistic policy of Section~\ref{sec:env_and_architecture}
We make the room 5 by 5 meters (halving each dimension) and divide the agent's maximum ego-motion (forward/backward $v_{fwd}$ and planar angular $v_{\theta}$) by two while keeping the maximum interaction distance $\delta = 2$ fixed.
The same $16$ objects are placed in $16$ environments, one per environment, as in the 1-object experiments of Section~\ref{sec:experiments}.
We find (Figure~\ref{fig:small_room}) that the same sorts of milestones (ego-motion learning, object attention, improved object dynamics prediction) emerge, with similar comparisons to baseline, only approximately 4 times as fast.
Interestingly, after some time, the world-model loss dipped, and we hypothesize (but due to computational constraints, did not run out sufficiently long, given this 4x heuristic) that we would see this behavior in our main setup, as well.

\subsection{Object frequency breakdown.} \label{sec:object_frequencies}

To measure how learned object attention depends on shape, we modify our training procedure, assigning each of our 16 environments a unique shape --- each environment is assigned a single shape that it uses for each reset, so that at all points in training, each shape is in exactly one environment. The environment and agent parameters are as described in Section~\ref{sec:env_stability}, differing in environment dimensions and agent speeds from our main experimental setups. We then measure object play frequency broken down by object (Figure~\ref{fig:objectablation}). Note the heterogeneity --- while most objects have similar play frequency graphs, others have inconsistent play frequency. This suggests that the control problem of finding an object, and keeping it in view, is not learned with equal success across objects.

\begin{figure}
  \begin{center}
    \includegraphics[width=\textwidth]{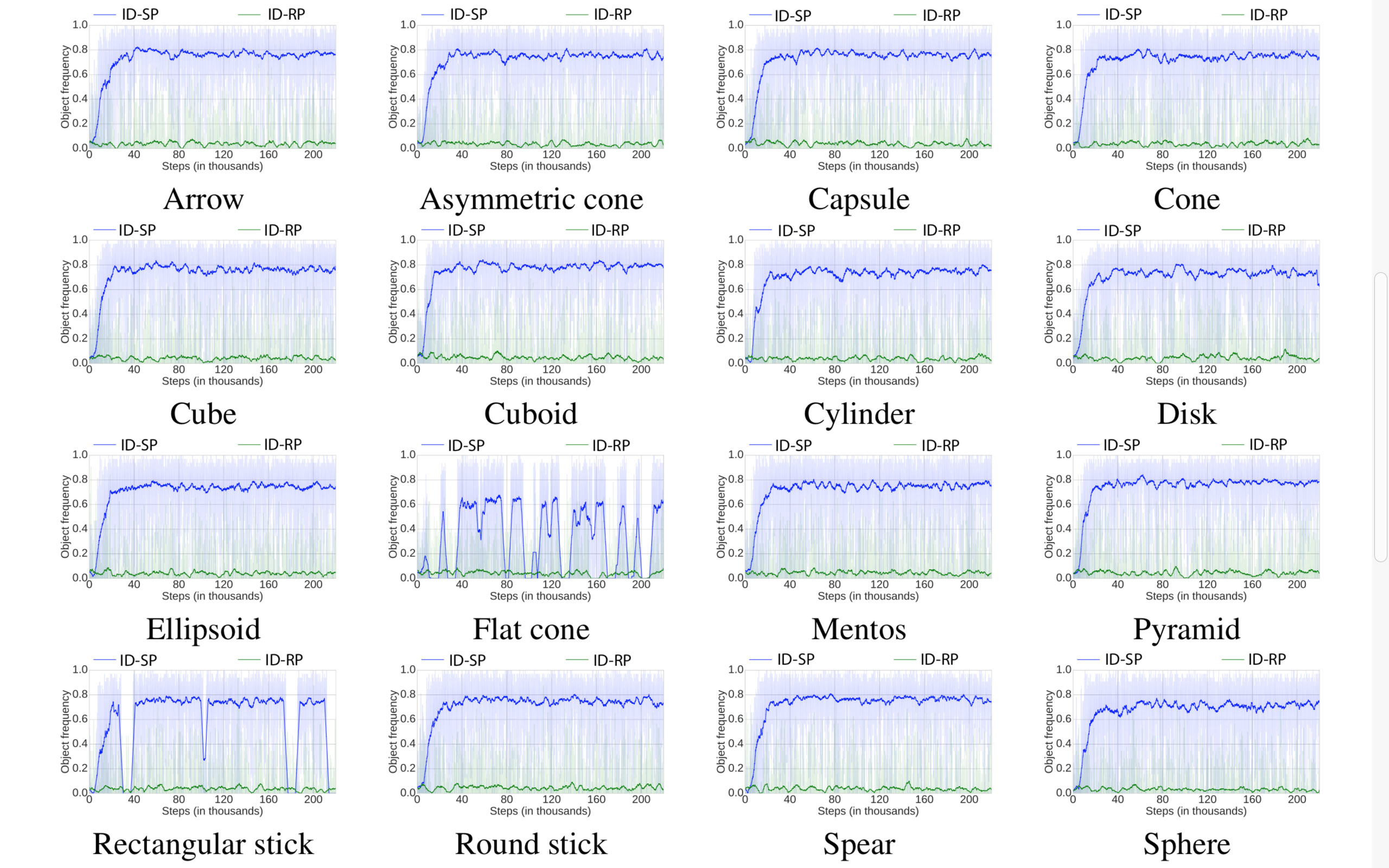}
  \end{center}
  \label{fig:objectablation}
  \caption{Object in view frequency across all objects and for each of the tested 16 objects individually.}
\end{figure}

\end{document}